\documentclass{article} 
\usepackage{nips13submit_e,times}

\usepackage{graphicx} 
\usepackage{subfigure}
\usepackage{amssymb}
\usepackage{amsfonts}
\usepackage{amsmath}
\usepackage{amsthm}
\usepackage{latexsym}
\usepackage{url}
\usepackage{enumerate}
\usepackage{wrapfig}

\usepackage{algorithm}
\usepackage{algorithmic}

\usepackage{picinpar}
\usepackage{pgf}

\newcommand{\tr}{\mathrm{tr}}
\newcommand{\cut}{\mathrm{Cut}}
\newcommand{\hess}{\mathrm{Hess}}
\newcommand{\HS}{\mathrm{HS}}
\newcommand{\cM}{\mathcal{M}}

\makeatletter\def\@captype{figure}\makeatother

\newtheorem{Them}{Theorem}[section]
\newtheorem{Def}{Definition}[section]

\newtheorem*{Exam*}{Example}
\newtheorem*{Rem*}{Remark}

\title{Geodesic Distance Function Learning via \\heat flow on Vector Fields}

\author{
Binbin Lin$^\dagger$ \hspace{0.5cm}  Ji Yang$^\ddagger$  \hspace{0.5cm}  Xiaofei He$^\ddagger$  \hspace{0.5cm}  Jieping Ye$^\dagger$    \\
$^\dagger$Center for Evolutionary Medicine and Informatics, \\
Arizona State University, Tempe AZ 85287, USA \\
$^\ddagger$State Key Lab of CAD$\&$CG, College of Computer Science,\\
Zhejiang University, Hangzhou 310058, China 
}

\date{\today}

\nipsfinalcopy 

\begin{document}

\maketitle

\begin{abstract}
Learning a distance function or metric on a given data manifold is of great importance in machine learning and pattern recognition. Many of the previous works first embed the manifold to Euclidean space and then learn the distance function. However, such a scheme might not faithfully preserve the distance function if the original manifold is not Euclidean. Note that the distance function on a manifold can always be well-defined. In this paper, we propose to learn the distance function directly on the manifold without embedding. We first provide a theoretical characterization of the distance function by its gradient field. Based on our theoretical analysis, we propose to first learn the gradient field of the distance function and then learn the distance function itself. Specifically, we set the gradient field of a local distance function as an initial vector field. Then we transport it to the whole manifold via heat flow on vector fields. Finally, the geodesic distance function can be obtained by requiring its gradient field to be close to the normalized vector field. Experimental results on both synthetic and real data demonstrate the effectiveness of our proposed algorithm.
\end{abstract}

\section{Introduction}

Learning a distance function or metric on a given data manifold is of great importance in machine learning and pattern recognition. The goal of \emph{distance metric learning} on the manifold is to find a desired distance function $d(x, y)$ such that it provides a natural measure of the \emph{similarity} between two data points $x$ and $y$ on the manifold. It has been applied widely in many problems, such as information retrieval~\cite{McFeeL10}, classification and clustering~\cite{XingNJR02}. Depending on whether there is label information available, metric learning methods can be classified into two categories: supervised and unsupervised. In supervised learning, one often assumes that data points with the same label should have small distance, while data points with different labels should have large distance~\cite{XingNJR02, Weinberger06distance,Jin09RDML}. In this paper, we consider the problem of unsupervised distance metric learning.

Unsupervised manifold learning can be viewed as an alternative way of distance metric learning. It aims to find a map $F$ from the original high dimensional space to a lower dimensional Euclidean space such that the mapped Euclidean distance $d(F(x), F(y))$ preserves the original distance $d(x, y)$. The classical Principal Component Analysis (PCA,~\cite{PCA}) can be considered as linear manifold learning method in which the map $F$ is linear. The learned Euclidean distance after linear mapping is also referred to as Mahalanobis distance. Note that when the manifold is nonlinear, the Mahalanobis distance may fail to faithfully preserve the original distance.

The typical nonlinear manifold learning approaches include Isomap~\cite{Isomap}, Locally Linear Embedding (LLE,~\cite{LLE}), Laplacian Eigenmaps (LE,~\cite{eigenmap}), Hessian Eigenmaps (HLLE,~\cite{HLLE}), Maximum Variance Unfolding (MVU,~\cite{MVU}) and Diffusion Maps \cite{Coifman20065}. Both Isomap and HLLE try to preserve the original geodesic distance on the data manifold. Diffusion maps try to preserve diffusion distance on the manifold which reflects the connectivity of data. Coifman and Lafon~\cite{Coifman20065} also showed that both LLE and LE belong to the diffusion map framework which preserves the local structure of the manifold. MVU is proposed to learn a kernel eigenmap that preserves pairwise distances on the manifold. One problem of the existing manifold learning approaches is that there may not exist a distance preserving map $F$ such that $d(F(x), F(y)) = d(x, y)$ holds since the geometry and topology of the original manifold may be quite different from the Euclidean space. For example, there does not exist a distance preserving map between a sphere $S^2$ and a 2-dimensional plane.

In this paper, we assume the data lies approximately on a low-dimensional manifold embedded in Euclidean space. Our aim is to approximate the geodesic distance function on this manifold. The geodesic distance is a fundamental intrinsic distance on the manifold and many useful distances (e.g., the diffusion distance) are variations of the geodesic distance. There are several ways to characterize the geodesic distance due to its various definitions and properties. The most intuitive and direct characterization of the geodesic distance is by definition that it is the shortest path distance between two points (e.g.,~\cite{Isomap}). However, it is well known that computing pairwise shortest path distance is time consuming and it cannot handle the case when the manifold is not geodesically convex~\cite{HLLE}. A more convincing and efficient way to characterize the geodesic distance function is using partial differential equations (PDE). M{\'e}moli \emph{et al.}~\cite{Memoli2001730} proposes an iterated algorithm for solving the Hamilton-Jacobi equation $\| \nabla r \| =1$~\cite{mantegazza2003hamilton}, where $\nabla r$ represents the gradient field of the distance function. However, the fast marching part requires a grid of the same dimension as the ambient space which is impractical when the ambient dimension is very high.

Note that the tangent space dimension is equal to the manifold dimension~\cite{LEESmoothManifold} which is usually much smaller than the ambient dimension. One possible way to reduce the complexity of representing the gradient field $\nabla r$ is to use the local tangent space coordinates rather than the ambient space coordinates. Inspired by recent work on vector fields {\cite{Singer2011Vector, Lin:2011:PFR, lin:2013:pfe}} and heat flow on scalar fields  \cite{Crane:2013:GHN}, we propose to learn the geodesic distance function via the characterization of its gradient field and heat flow on vector fields. Specifically, we study the geodesic distance function $d(p, x)$ for a given base point $p$. Our theoretical analysis shows that if a function $r_p(x)$ is a local distance function around $p$, and its gradient field $\nabla r_p$ has unit norm or $\nabla r_p$ is parallel along geodesics passing through $p$, then $r_p(x)$ is the unique geodesic distance function $d(p, x)$. Based on our theoretical analysis, we propose a novel algorithm to first learn the gradient field of the distance function and then learn the distance function itself. Specifically, 
we set the gradient field of a local distance function around a given point as an initial vector field. Then we transport the initial local vector field to the whole manifold via heat flow on vector fields. By asymptotic analysis of the heat kernel, we show that the learned vector field is approximately parallel to the gradient field of the distance function at each point. Thus, the geodesic distance function can be obtained by requiring its gradient field to be close to the normalized vector field. The corresponding optimization problem involves sparse linear systems which can be solved efficiently. Moreover, the sparse linear systems can be easily extended to matrix form to learn the complete distance function $d(\cdot, \cdot)$. Both synthetic and real data experiments demonstrate the effectiveness of our proposed algorithm.

\section{Characterization of Distance Functions using Gradient Fields}\label{sec:distance-function-vector-field}

Let $(\mathcal{M},g)$ be a $d$-dimensional Riemannian manifold, where $g$ is a Riemannian \emph{metric tensor} on $\mathcal{M}$. The goal of \emph{distance metric learning} on the manifold is to find a desired distance function $d(x, y)$ such that it provides a natural measure for the \emph{similarity} between two data points $x$ and $y$ on the manifold. In this paper, we study a fundamental intrinsic distance function\footnote{A distance function $d(\cdot, \cdot)$ defined by its Riemannian metric $g$ is often called an intrinsic distance function.} - the geodesic distance function. Similar to many geometry textbooks~(e.g., \cite{Jost08Geometry, RiemannianGeometry}), we call it the distance function. In the following, we will briefly introduce the most relevant concepts. A detailed and much more technical introduction can be found in the appendix.

We first show how to assign a metric structure on the manifold. For each point $p$ on the manifold, a Riemannian metric tensor $g$ at $p$ is an inner product $g_p$ on each of the tangent space $T_{p}\mathcal{M}$ of $\mathcal{M}$. We define the norm of a tangent vector $v\in T_p\mathcal{M}$ as $\|v\| = \sqrt{g_p(v, v)}$. Let $[a,b]$ be a closed interval in $\mathbb{R}$, and $\gamma:[a,b]\rightarrow \mathcal{M}$ be a smooth curve. The \emph{length} of $\gamma$ can then be defined as
$
l(\gamma):= \int_{a}^b \|\frac{d\gamma}{dt}(t)\|dt.
$
The \emph{distance} between two points $p, q$ on the manifold $\mathcal{M}$ can be defined as:
\[
\begin{aligned}
d(p,q) := \inf \{ &l(\gamma): \gamma:[a,b]\rightarrow \mathcal{M}~ \mathrm{piecewise~ smooth}, \\
&\gamma(a) = p \mathrm{~and~}\gamma(b)=q  \}.
\end{aligned}
\]
We call $d(\cdot,\cdot)$ the \emph{distance function} and it satisfies the usual axioms of a metric, i.e., positivity, symmetry and triangle inequality~\cite{Jost08Geometry}. We study the distance function $d(p, \cdot)$ when $p$ is given.
\begin{Def}[Distance function based at a point]
Let $\mathcal{M}$ be a Riemannian manifold, and let $p$ be a point on the manifold $\mathcal{M}$. A \emph{distance function} on $\mathcal{M}$ based at $p$ is defined as $r_p(x) = d(p, x)$. For simplicity, we might write $r(\cdot)$ instead of $r_p(\cdot)$.
\end{Def}

\begin{figure}[t]\centering
    \subfigure[$r_p$]{\includegraphics[width=.32\linewidth]{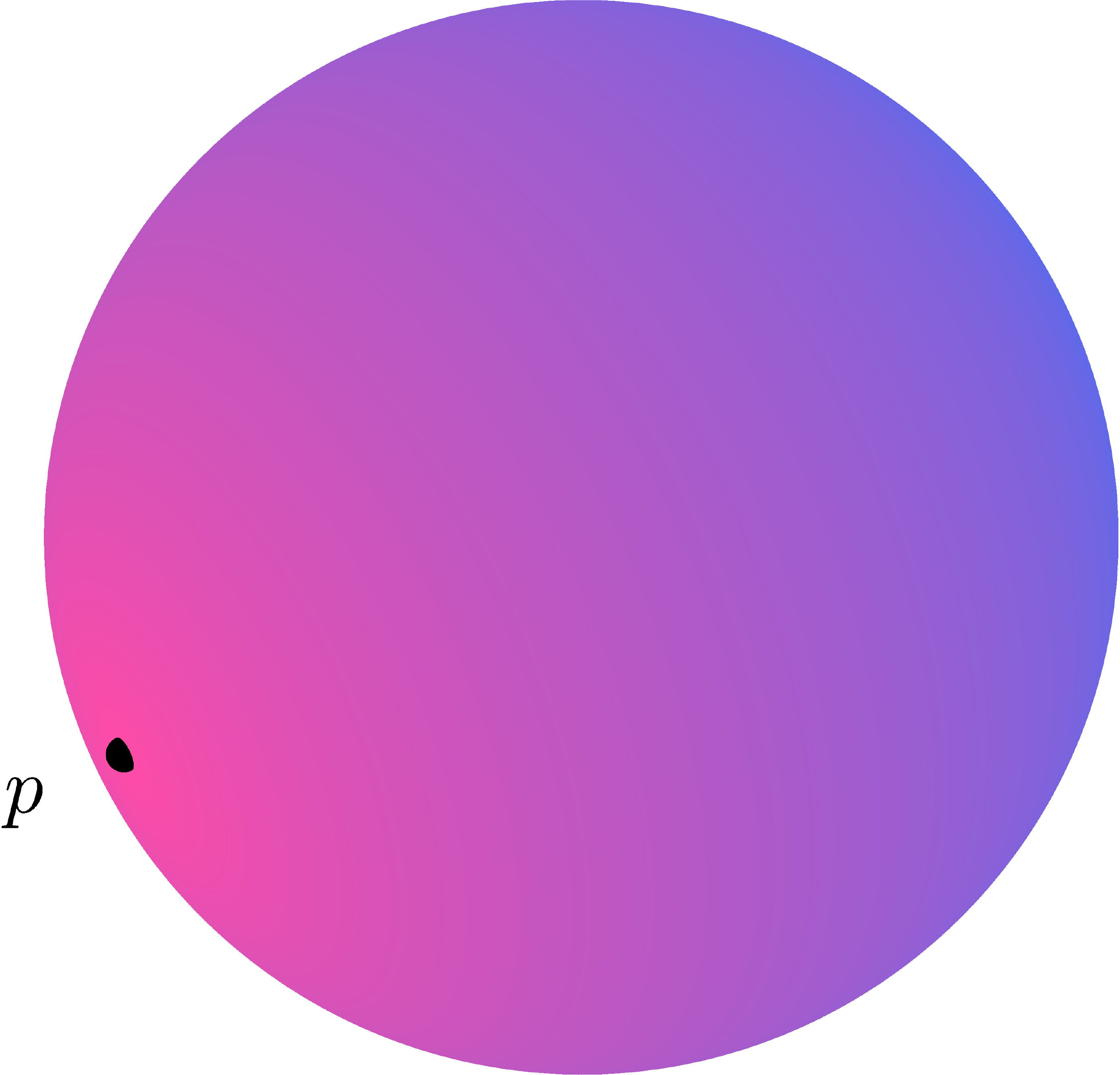}}
    \subfigure[$\nabla r_p$]{\includegraphics[width=.32\linewidth]{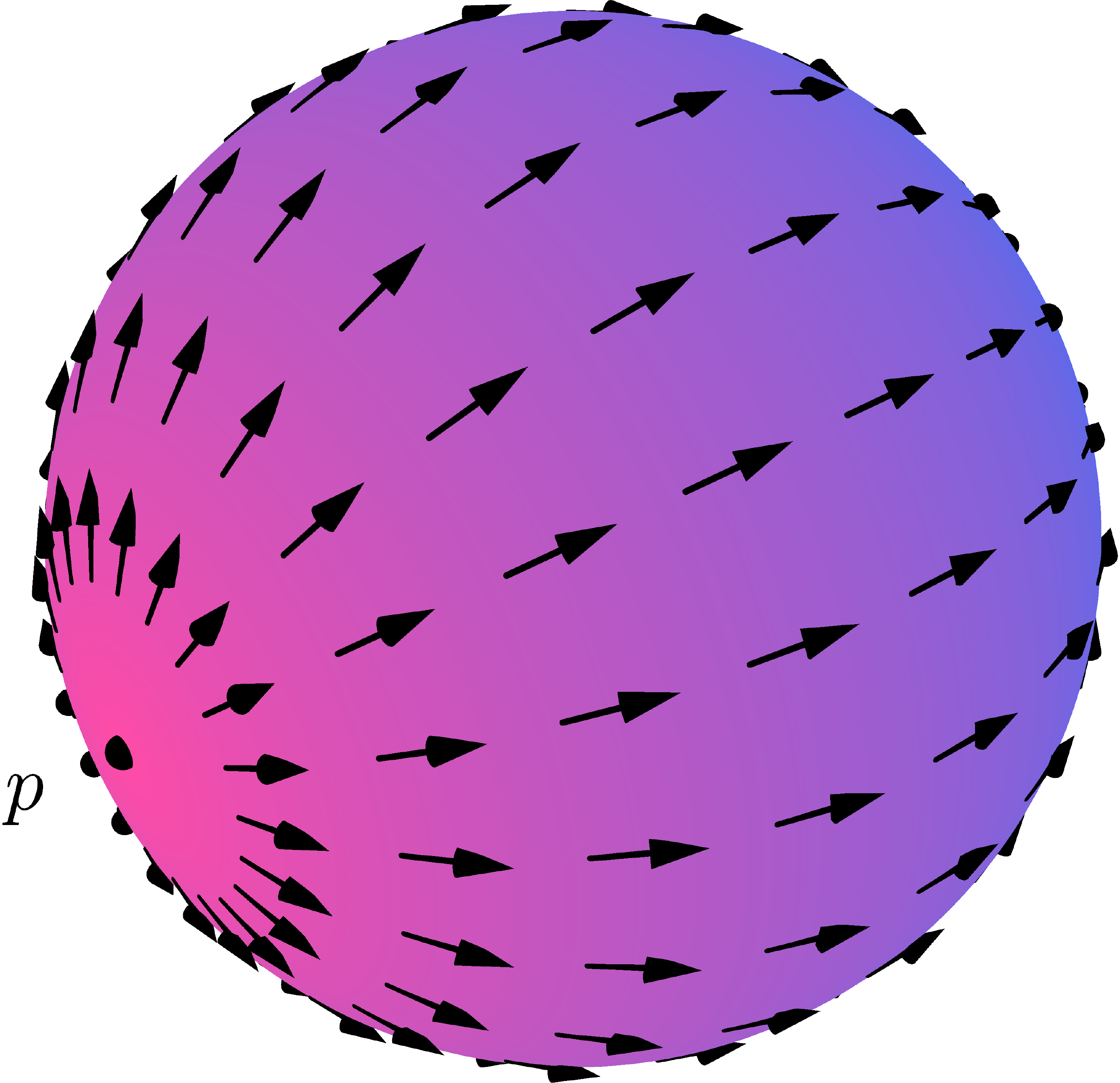}}
    \subfigure[Geodesics]{\includegraphics[width=.32\linewidth]{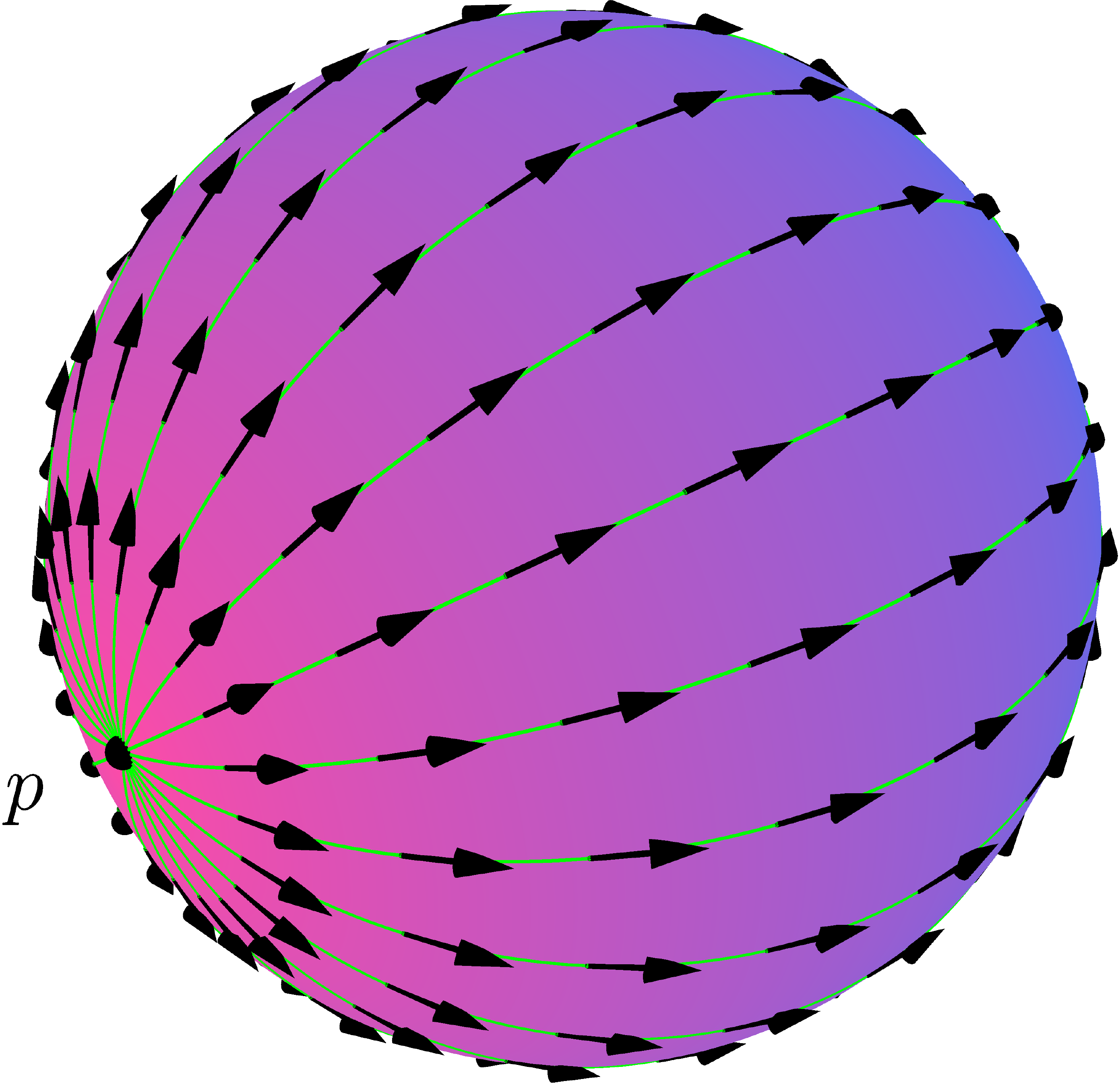}}
    \caption{(a) shows the distance function $r_p(x)$ on the sphere, where the base point $p$ is marked in black. Different colors indicate different distance values. (b) shows the gradient field $\nabla r_p$. (c) shows the geodesics passing through $p$ which are denoted by the green lines.
    }
    \label{fig:gradientfield-geodesic}
\vskip -0.2in
\end{figure}

\begin{Def}[Geodesic,~\cite{RiemannianGeometry}]
Let $\gamma: [a, b]\rightarrow \mathcal{M}, t\mapsto \gamma(t)$ be a smooth curve. $\gamma$ is called a \emph{geodesic} if ${\gamma'}(t)$ is parallel along $\gamma$, i.e., $ \nabla_{\gamma'(t)}\gamma'(t)=0$ for all $t\in [a, b]$.
\end{Def}
Here $\nabla$ is the covariant derivative on the manifold which measures the change of vector fields. A geodesic can be viewed as a \emph{curved straight line} on the curved manifold. The geodesics and the distance function is related as follows:
\begin{Them}[\cite{RiemannianGeometry}]\label{Them:segments_are_geodesics}
If $\gamma$ is a local minimum for $\inf l(\gamma)$ with fixed end points, then $\gamma$ is a geodesic.
\end{Them}

In the following, we characterize the distance function $r_p(x)$ by using its gradient field $\nabla r$. A vector field $X$ on the manifold is called a \emph{gradient field} if there exists a function $f$ on the manifold such that $X = \nabla f$ holds. Therefore, gradient fields are one kind of vector fields. Interestingly, we can precisely characterize the distance function based at a point by its gradient field. For simplicity, let $\partial_r$ denote the gradient field $\nabla r$. Then we have the following result.
\begin{Them}\label{Them:distance_function_vector_field_equivalence}
Let $\mathcal{M}$ be a complete manifold. A continuous function $r: \mathcal{M}\rightarrow \mathbb{R}$ is a distance function on $\mathcal{M}$ based at $p$ if and only if (a) $r(x) = \|\exp_p^{-1}(x)\|$ holds for a neighborhood of $p$; (b) $\nabla_{\partial r} \partial r = 0$ holds on the manifold $\mathcal{M}$ except for $p\cup \mathrm{Cut}(p)$.
\end{Them}
Here $\exp_p$ is the exponential map at $p$ and $\mathrm{Cut}(p)$ is the cut locus of $p$. The detailed definitions of the exponential map and the cut locus can be found in the appendix. Condition (a) states that locally $r(x)$ is a Euclidean distance function in the exponential coordinates. Combining condition (b) which states that the integral curves of $\partial_r$ are all geodesics, we assert that $r$ is a global distance function. As can be seen from Fig.~\ref{fig:gradientfield-geodesic}(c), the gradient field of the distance function is parallel along the geodesics passing through $p$. It might be worth noting that condition (a) cannot be replaced by a weaker condition $r(p) = 0$ which is often used in PDE. A simple counter-example would be the function $r_p(x) = x$ defined on $\mathcal{M} = \mathbb{R}$ with $p=0$. $r_p(x)$ satisfies $r_p(0) = 0$ and $\nabla_{\partial_r} \partial_r = 0$ holds for all $x$. However, it is not a distance function since it does not satisfy the positivity condition.

The second order condition $\nabla_{\partial_r} \partial_r = 0$ can be replaced by a first order condition $\| \partial_r \| =1$.
\begin{Them}\label{Them:distance_function_unit_norm}
Let $\mathcal{M}$ be a complete manifold. A continuous function $r: \mathcal{M}\rightarrow \mathbb{R}$ is a distance function on $\mathcal{M}$ based at $p$ if and only if (a) $r(x) = \|\exp_p^{-1}(x)\|$ holds for a neighborhood of $p$; (b) $\| \partial_r \| =1$ holds on the manifold $\mathcal{M}$ except for $p\cup \mathrm{Cut}(p)$.
\end{Them}

A detailed proof of Theorems~\ref{Them:distance_function_vector_field_equivalence} and \ref{Them:distance_function_unit_norm} can be found in the appendix. We visualize the relationship among the distance function, the gradient field of the distance function and geodesics in Fig.~\ref{fig:gradientfield-geodesic}. It can be seen from the figure that: (1) the gradient field of the distance function is parallel along geodesics passing through $p$; (2) the gradient field of the distance function has unit norm almost everywhere except for $p$ and its cut locus which is the antipodal point of $p$.

\section{Geodesic Distance Function Learning}
We show in the last section that the distance function can be characterized by its gradient field. Based on our theoretical analysis, we propose to first learn the gradient field of the distance function and then learn the distance function itself. 

\begin{figure}[t]\small\centering
    \subfigure[Initial $V^0$]{\includegraphics[width=.22\linewidth]{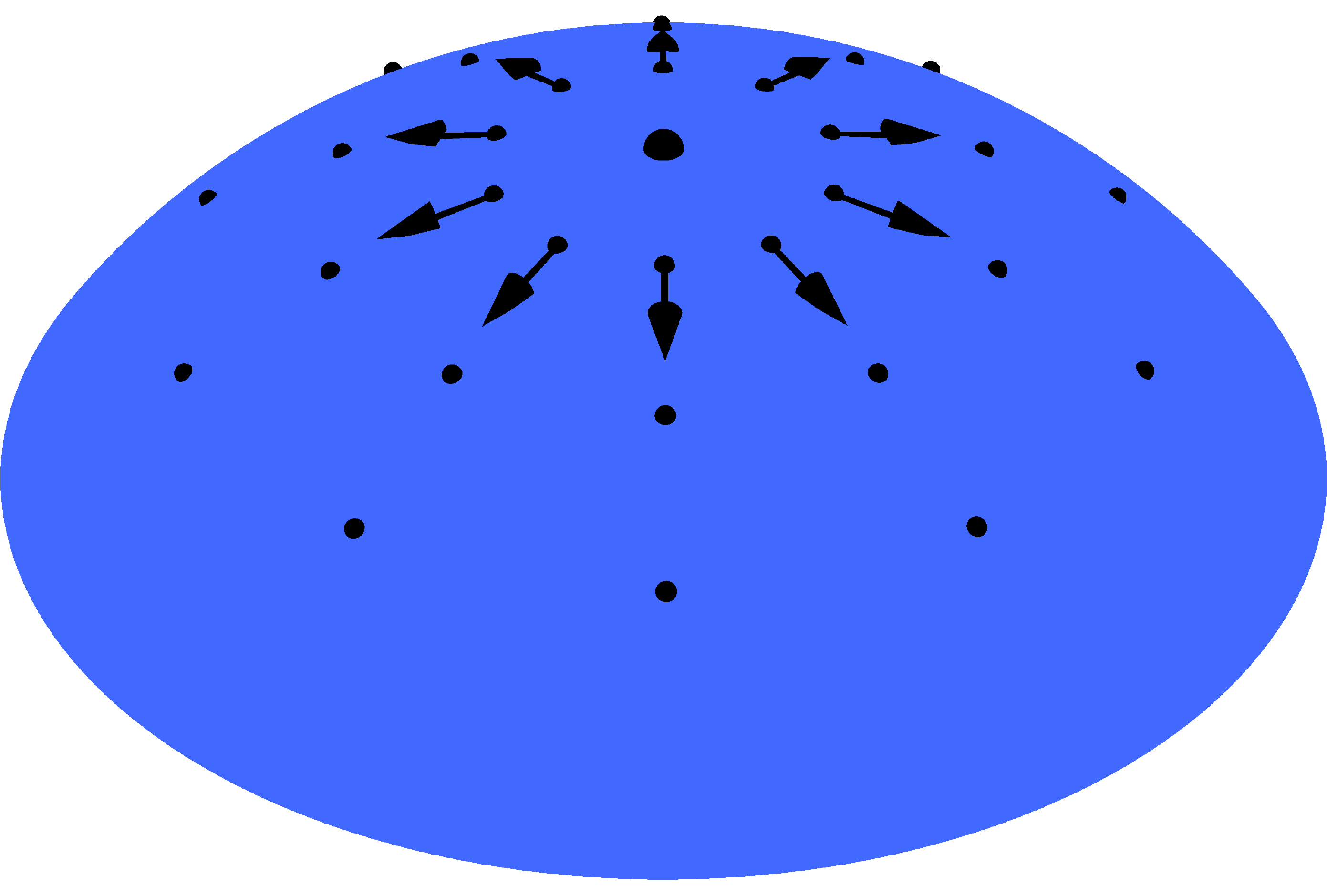}}\hspace{0.05in}
    \subfigure[Heat flow $V$]{\includegraphics[width=.22\linewidth]{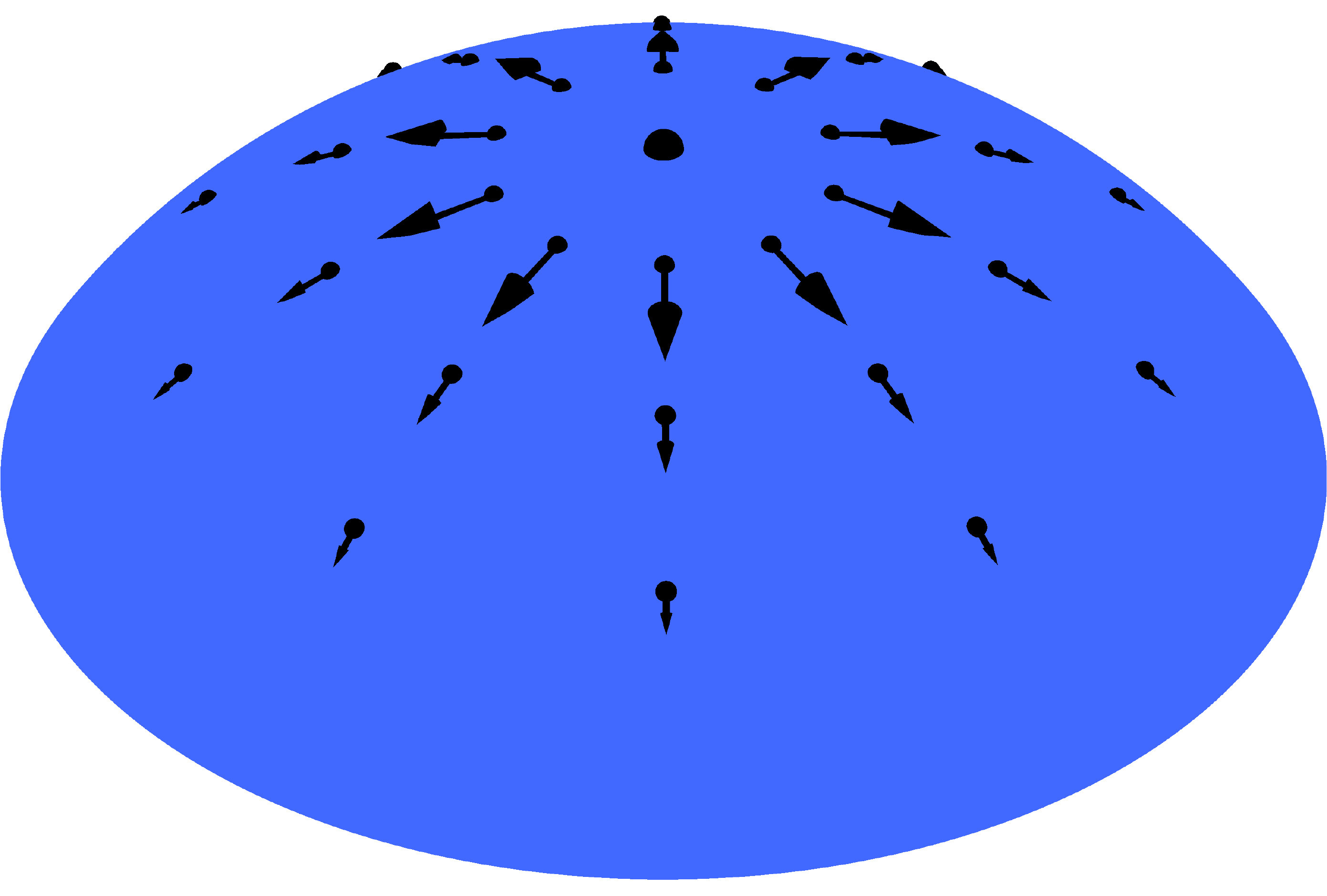}}\hspace{0.05in}
    \subfigure[Normalized $\hat{V}$]{\includegraphics[width=.22\linewidth]{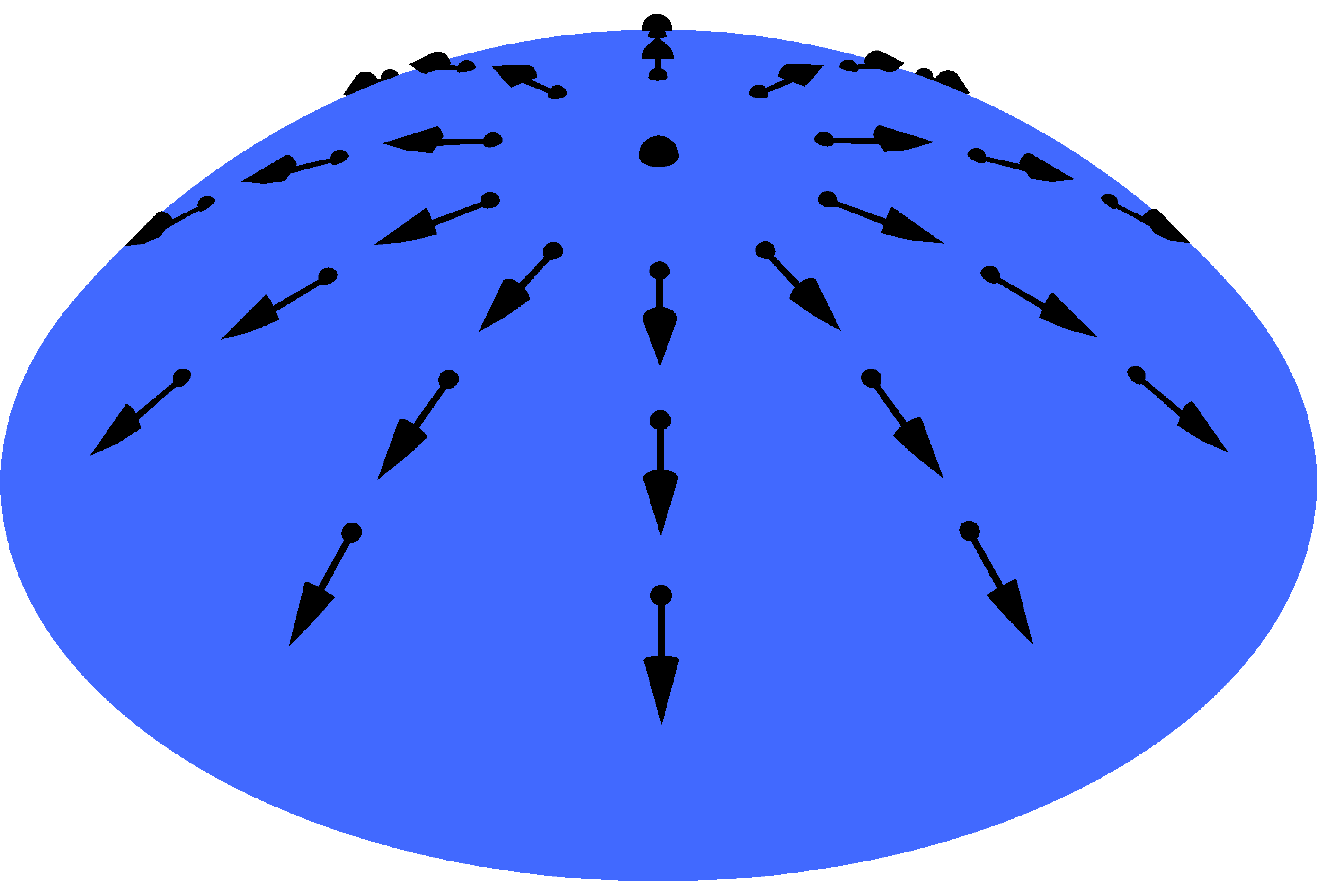}}\hspace{0.05in}
    \subfigure[Distance function $f$]{\includegraphics[width=.22\linewidth]{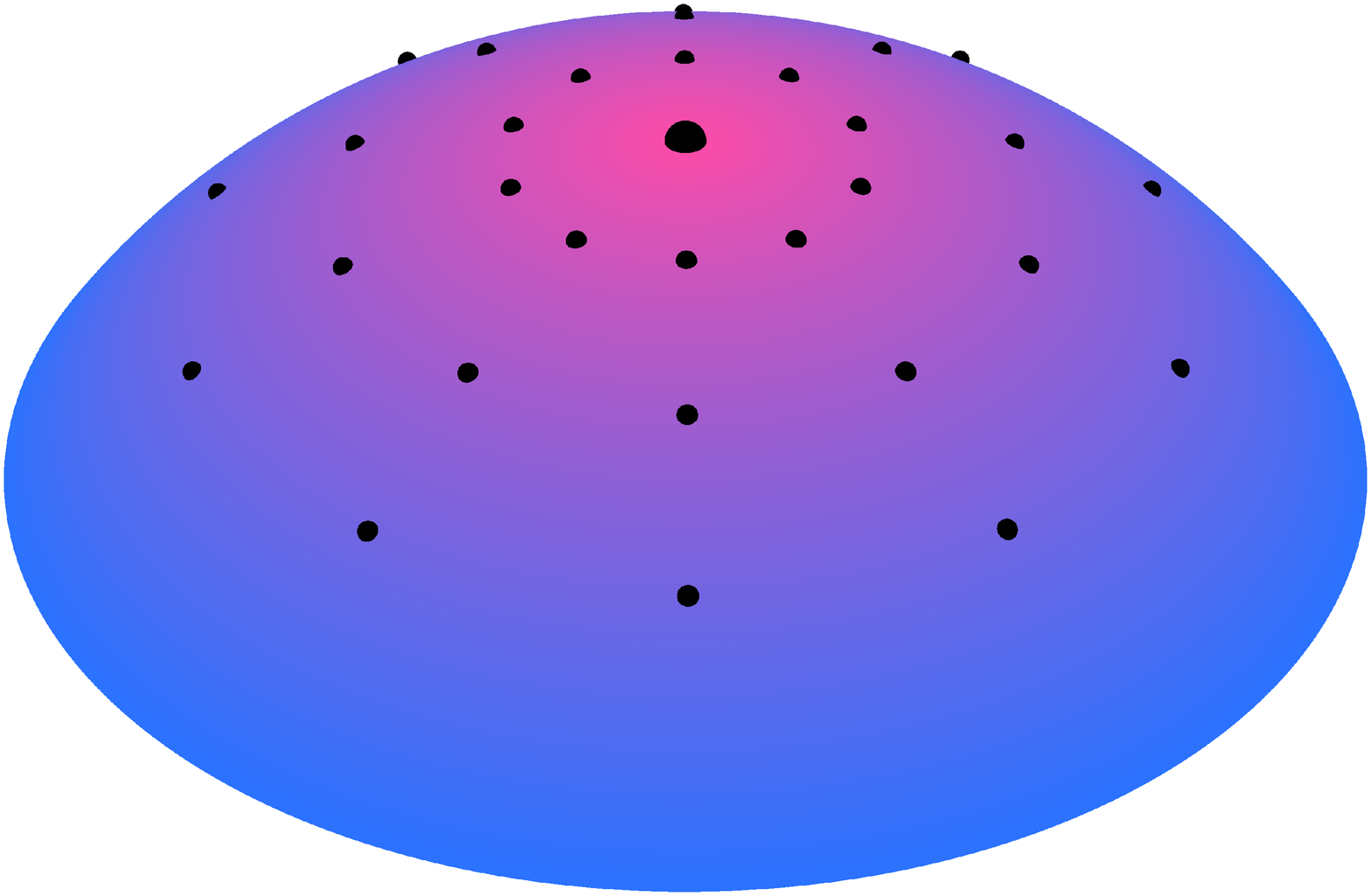}}
    \vspace{-0.1in}
    \caption{Algorithm overview. The base point is on the top of the manifold. (a) shows the initial vector field $V^0$. (b) shows the vector field $V$ after transporting $V^0$ to the whole manifold using heat flow on vector fields. (c) shows the normalized vector field $\hat{V}$ of $V$. (d) shows the final distance function learned via requiring its gradient field to be close to $\hat{V}$, where the red color indicates small distance function values and the blue color indicates large distance function values.}
    \label{fig:algorithm-overview}
    \vspace{-0.1in}
\end{figure}

\subsection{Geodesic Distance Learning}\label{sec:algorithm}
Let $(\mathcal{M},g)$ be a $d$-dimensional Riemannian manifold embedded in a much higher dimensional Euclidean space $\mathbb{R}^m$, where $g$ is a Riemannian metric tensor on $\mathcal{M}$. Given a point $p$ on the manifold, we aim to learn the distance function $f_p(x) = d(p, x)$. Let $U_{\epsilon} :=\{ x: d(p, x) \leq \epsilon \} \subset \mathcal{M}$ be a geodesic ball around $p$ and let $f^0$ denote a local distance function on $U$. That is, $f^0(x) = d(p, x)$ if $p\in U$ and $0$ otherwise. Let $V^0$ denote the gradient field of $f^0$, i.e., $V^0 = \nabla f^0$. Now we are ready to summarize our Geodesic Distance Learning (GDL) algorithm as follows:
\begin{itemize}
  \item Learn a vector field $V$ by transporting $V^0$ to the whole manifold using heat flow:
\begin{equation}\label{eq:vf-continuous-obj}
\min_{V} E(V) := \int_{\mathcal{M}} \| V - V^0\|^2 dx + t \int_{\mathcal{M}} \| \nabla V\|_{\mathrm{HS}}^2 dx,
\end{equation}
where $\|\cdot\|_{\HS}$ denotes the Hilbert-Schmidt tensor norm \cite{Defant1993tensor} and $t>0$ is a parameter.
  \item Learn a normalized vector field $\hat{V}$ via normalizing $V$ at each point $x$: set $\hat{V}_x = V_x / \| V_x \|$ when $x\neq p$ and set $\hat{V}_x  = 0$ when $x=p$. Here $V_x$ denotes the tangent vector at $x$ of $V$.
  \item Learn the distance function $f$ via solving the following equation:
\begin{equation}\label{eq:final-function}
\min_f \Phi(f) := \int_{\mathcal{M}} \| \nabla f - \hat{V} \|^2 dx, ~~ \mathrm{s.t.  } f(p) =0.
\end{equation}
\end{itemize}
The above algorithmic steps are illustrated in Fig.~\ref{fig:algorithm-overview}.

The theoretical justification of the above algorithm is given in the appendix. Our analysis indicates that solving Eq.~\eqref{eq:vf-continuous-obj} is equivalent to transporting the initial vector field to the whole manifold via heat flow on vector fields. By asymptotic analysis of the heat kernel, the learned vector field is approximately parallel to the gradient field of the distance function at each point. Thus, the gradient field of the distance function can be obtained via normalization. Finally, the geodesic distance function can be obtained by requiring its gradient field to be close to the normalized vector field. Our analysis also indicate the factors of controlling the quality of the approximation. It mainly relies on two factors: the distance to the query and the cut locus of the query. If the data point is not in the cut locus of the query, the smaller the distance between the data point and the query is, the better the approximation would be. If the data point is in the cut locus, the approximation might fail since the vector field around the cut locus varies dramatically. Note that the measure of the cut locus is zero, thus the approximation would fail only in a zero measure set.

\subsection{Implementation}\label{sec:implementation}

Given $n$ data points $x_i, i=1,\ldots, n$, on the $d$-dimensional manifold $\mathcal{M}$ where $\mathcal{M}$ is embedded in the high dimensional Euclidean space $\mathbb{R}^m$. Let $x_q$ denote the base point. We aim to learn the distance function $f: \mathcal{M} \rightarrow \mathbb{R}$ based at $x_q$, i.e., $f(x_i) = d(x_q, x_i)$, $i=1,\ldots, n$.

We first construct an undirected nearest neighbour graph by either $\epsilon$-neighbourhood or $k$ nearest neighbours. It might be worth noting for a $k$-nn graph that the degree of a vertex will typically be larger than $k$ since $k$ nearest neighbour relationships are not symmetrical. Let $x_i \sim x_j$ denote that $x_i$ and $x_j$ are neighbors. Let $w_{ij}$ denote the weight which can be approximated by the heat kernel weight or the simple 0-1 weight. For each point $x_i$, we estimate its tangent space $T_{x_i}\mathcal{M}$ by performing PCA on its neighborhood. Before performing PCA, we mean-shift the neighbor vectors using their true mean. Let $T_i\in\mathbb{R}^{m\times d}$ denote the matrix whose columns are constituted by the $d$ principal components. Let $V$ be a vector field on the manifold. For each point $x_i$, let $V_{x_i}$ denote the tangent vector at $x_i$. Recall from Definition~\ref{def:ts} in the appendix that each tangent vector $V_{x_i}$ should be in the corresponding tangent space $T_{x_i}\mathcal{M}$, we can represent $V_{x_i}$ as $V_{x_i}=T_iv_i$, where $v_i\in\mathbb{R}^d$. We will abuse the notation $f$ to denote the vector $f = (f(x_1),\ldots,f(x_{n}))^T \in \mathbb{R}^n$ and use $V$ to denote the vector $V=\left( {v_1}^T, \ldots, {v_{n}}^T \right)^T \in \mathbb{R}^{dn}$. We propose to first learn $V$ and then learn $f$.

Set an initial vector field $V^0$ as follows:
\begin{equation}\label{eq:set-initial-v0-dis}
v^0_j = \left\{
\begin{aligned}
\frac{T_j^T(x_j-x_q)}{\|T_jT_j^T(x_j-x_q)\|} , &\quad \mathrm{ if } ~j\sim q\\
0, &\quad \mathrm{ otherwise }
\end{aligned}
\right.
\end{equation}
Note that the vector $T_j^T(x_j-x_q)/{\|T_jT_j^T(x_j-x_q)\|}$ is a unit vector at $x_j$ pointing outward from the base point $x_q$ (please see Fig.~\ref{fig:algorithm-overview}(a)). Following~\cite{Lin:2011:PFR}, the discrete form of our objective functions can be given as follows:
\begin{equation}\label{eq:discrete-obj}
\begin{aligned}
 E(V) &= V^T  V - 2 {V^0}^T  {V} + {V^0}^T  {V^0} + t V^T B V,\\
 \Phi(f) &= 2f^T L f + \hat{V}^T G \hat{V} - 2 \hat{V}^T C f,
\end{aligned}
\end{equation}
where $L$ is the graph Laplacian matrix~\cite{Chung}, $B$ is a $dn\times dn$ block matrix, $G$ is a $dn\times dn$ block diagonal matrix and $C$ is a $dn\times n$ block matrix. Let $B_{ij}$ $(i\neq j)$ denote the $ij$-th $d\times d$ block, $G_{ii}$ denote the $i$-th $d\times d$ diagonal block of $G$, and $C_i$ denote the $i$-th $d\times n$ block of $C$. We have: $B_{ii}=\sum_{j \sim i}w_{ij}(Q_{ij}Q_{ij}^T+I)$, $B_{ij}= -2w_{ij}Q_{ij}$, $G_{ii} = \sum_{j\sim i} w_{ij}T_i^{T} (x_j-x_i)(x_j-x_i)^{T} T_i$, and $C_i = \sum_{j\sim i} w_{ij} T_i^{T}(x_j-x_i)s_{ij}^{T}$, where $ Q_{ij}= T_i^T T_j$ and $s_{ij}\in \mathbb{R}^n$ is a selection vector of all zero elements except for the $i$-th element being $-1$ and the $j$-th element being $1$. The matrix $Q_{ij}$ transports from the tangent space $T_{x_j}\mathcal{M}$ to $T_{x_i}\mathcal{M}$ which approximates the parallel transport from $x_j$ to $x_i$. It might be worth noting that one can also approximate the parallel transport by solving a singular value decomposition problem~\cite{Singer2011Vector}. The block matrix $B$ provides a discrete approximation of the connection Laplacian operator, which is a symmetric and positive semi-definite matrix.

\begin{algorithm}
  \caption{GDL (Geodesic Distance Learning)}\label{alg:gdl-heat}
  \begin{algorithmic}
    \REQUIRE Data sample $X=(x_1,\ldots,x_n)\in\mathbb{R}^{m\times n}$ and a base point $x_q$, $1 \leq q \leq n$.
    \ENSURE $f=(f_1,\ldots,f_n)\in\mathbb{R}^{n}$
    \FOR{$i=1$ to $n$} \STATE Compute the tangent space coordinates $T_i \in \mathbb{R}^{m\times d}$ by using PCA
    \ENDFOR
    \vspace{5pt}
    \STATE Set an initial vector field $V^0$ via Eq.~\eqref{eq:set-initial-v0-dis} and construct sparse block matrices $B$ and $C$ 
    \STATE Solve $(I + t B) V = V^0$ to obtain $V$
    \STATE Normalize each vector in $V$ to obtain $\hat{V}$
    \STATE Solve $2L f = C^T \hat{V}$ to obtain $f$
    \RETURN $f$
  \end{algorithmic}
\end{algorithm}

Now we give our algorithm in the discrete setting. By taking derivatives of $E(V)$ with respect to $V$, $V$ can be obtained via the following sparse linear system:
\begin{equation}
(I + t B) V = V^0.
\label{eq:discrete-vector-field-heat-solution}
\end{equation}
Then we learn a normalized vector field $\hat{V}$ via normalizing $V$ at each point: $\hat{v_i} = v_i / \| v_i \|$ if $i\neq q$ and $\hat{v_i} = 0$ if $i=q$. The final distance function can be obtained via taking derivatives of $\Phi(f)$ with respect to $f$:
\begin{equation}
\label{eq:discrete-distanc-function-solution}
2 L f = C^T \hat{V},
\end{equation}
where we restrict $f_q = 0$ when solving Eq.~\eqref{eq:discrete-distanc-function-solution}. A direct way is to plug the constraint $f_q = 0$ into Eq.~\eqref{eq:discrete-distanc-function-solution}. It is equivalent to removing the $q$-th column of $L$ and the $q$-th element of $f$ on the left hand side of Eq.~\eqref{eq:discrete-distanc-function-solution}. For each point $x_q$, we have corresponding vectors $V$, $V^0$,$\hat{V}$ and$f$. If $x_q$ varies, we can write $V$, $V^0$, $\hat{V}$ and $f$ in matrix form where each column is a vector field or a distance function. Then the complete distance function $d(\cdot, \cdot)$ can be obtained via solving the corresponding matrix form linear systems of Eq.~\eqref{eq:discrete-vector-field-heat-solution} and Eq.~\eqref{eq:discrete-distanc-function-solution}. We summarize our algorithm in Algorithm \ref{alg:gdl-heat}.

\subsection{Computation Complexity Analysis}

The computational complexity of our proposed Geodesic Distance Learning (GDL) algorithm is dominated by three parts: searching for $k$-nearest neighbors, computing local tangent spaces, computing $Q_{ij}$ and solving the sparse linear system Eq.~\eqref{eq:discrete-vector-field-heat-solution}. For the $k$ nearest neighbor search, the complexity is $O((m+k)n^2)$, where $O(mn^2)$ is the complexity of computing the distance between any two data points, and $O(kn^2)$ is the complexity of finding the $k$ nearest neighbors for all the data points. The complexity for local PCA is $O(mk^2)$. Therefore, the complexity for computing the local tangent space for all the data points is $O(m n k^2 )$. Note that the matrix $B$ is not a dense matrix but a very sparse block matrix with at most $kn$ non-zero $d\times d$ blocks. Therefore the computation complexity of computing all $Q_{ij}$'s is $O(knmd^2)$. We use LSQR package\footnote{\url{http://www.stanford.edu/group/SOL/software/lsqr.html}} to solve Eq. ~\eqref{eq:discrete-vector-field-heat-solution}. It has a complexity of $O(Iknd^2)$, where $I$ is the number of iterations. In summary, the overall computational cost for one base point is
$
O((m+k)n^2 + mndk + kmnd^2 + Iknd^2).
$
For $p$ base points, the extra cost is to solve Eq.~\eqref{eq:discrete-vector-field-heat-solution} by adding $p-1$ columns which has a complexity of $O(pIknd^2)$. Empirically, the manifold dimension $d$ and the number of nearest neighbors $k$ are usually much smaller than the ambient dimension $m$ and the number of data points $n$. So the total computational cost for $p$ base points could be $O(mn^2 + pIn)$. There are several ways to further reduce the computational complexity. One way is to select anchor points and construct the graph using these anchor points. Another possible way is to learn the distance functions of nearby points simultaneously.

\subsection{Related Work and Discussion}
Our approach is based on the idea of vector field regularization which is similar to Vector Diffusion Maps (VDM,~\cite{Singer2011Vector}). Both methods employ vector fields to discover the geometry of the manifold. However, VDM and our approach differ in several key aspects: Firstly, they solve different problems. VDM tries to preserve the vector diffusion distance by dimensionality reduction while we try to learn the geodesic distance function directly on the manifold. It is worth noting that the vector diffusion distance is a variation of the geodesic distance. Secondly, they use different approximation methods. VDM approximates the parallel transport by learning an orthogonal transformation and we simply use projection adopted from \cite{Lin:2011:PFR}. GDL can also be regarded as a generalization of the heat method \cite{Crane:2013:GHN} on scalar fields. Both methods employ heat flow to obtain the gradient field of distance function. The algorithm proposed in \cite{Crane:2013:GHN} first learns a scalar field by heat flow on scalar fields and then learns the desired vector field by evaluating the gradient field of the obtained scalar field. Our method tries to learn the desired vector field directly by heat flow on vector fields. Note that the scalar field is zero order and the vector field is first order. It is expected that the first order approximation of the vector field might be more effective for high dimensional data.

There are several interesting future directions suggested in this work. One is to generalize the theory and algorithm in this paper to the multi-manifold case. The main challenge is that how to transport an initial vector field from one manifold to other manifolds. One feasible idea is to transport the vector from one manifold to another using the parallel transport in the ambient space since each tangent vector on the manifold is also a vector in the ambient space. Another direction is to estimate the true underlying manifold structure of the data despite the noise, e.g., the manifold dimension. We employ the tangent space structure to model the manifold and each tangent space is estimated by performing PCA. Note that the dimension of the manifold equals to the dimension of the tangent space. Therefore, if we can combine the work of PCA with noisy data and our framework, it might provide new methods and perspectives to the manifold dimension estimation problem. The third direction is to develop the machine learning theory and design efficient algorithms using heat flow on vector fields as well as other general partial differential equations.

\begin{figure}[t]\small
\begin{center}
\subfigure[Ground truth]{\includegraphics[width=.23\textwidth]{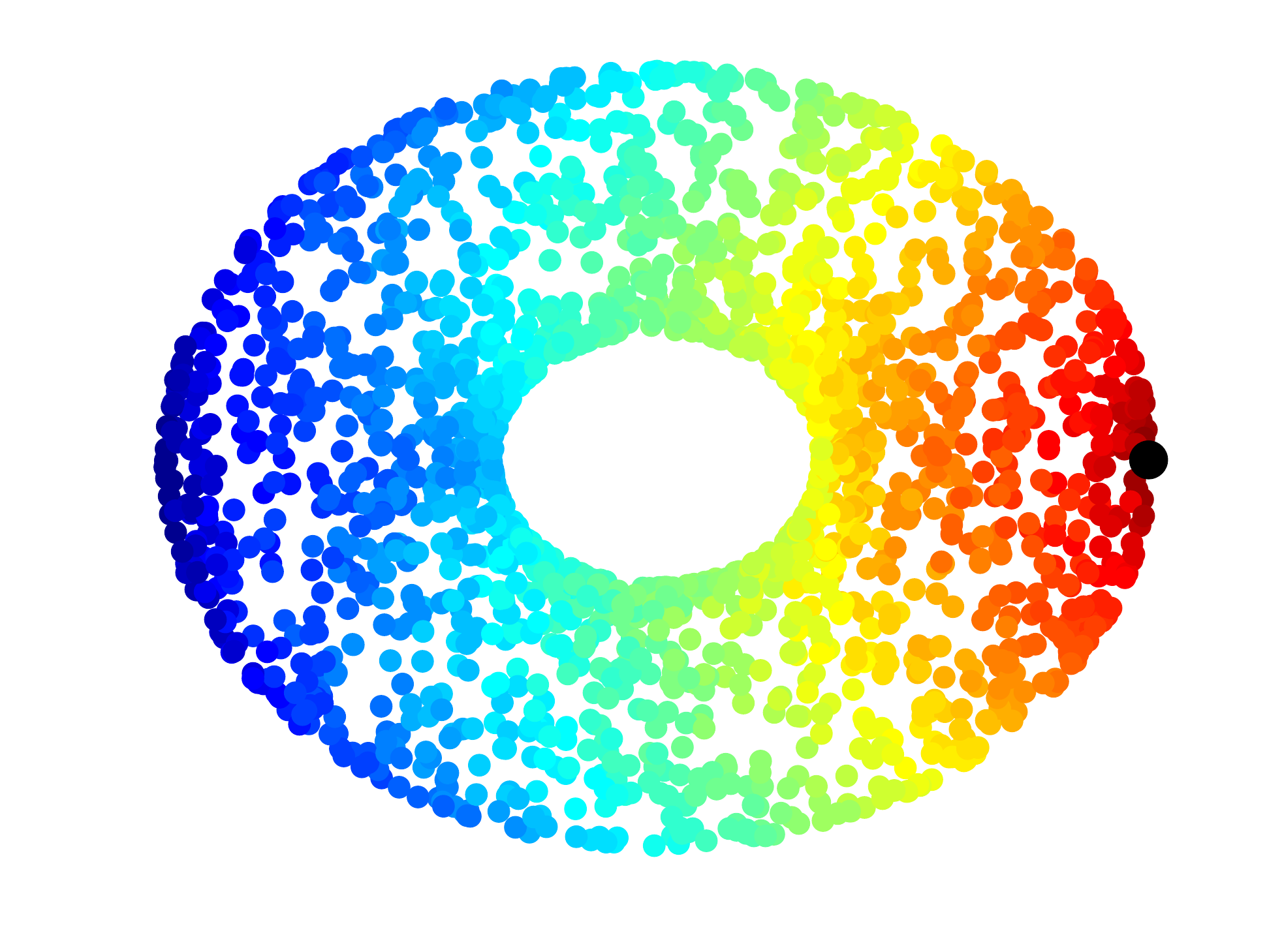}}
\subfigure[GDL (0.02)]{\includegraphics[width=.23\textwidth]{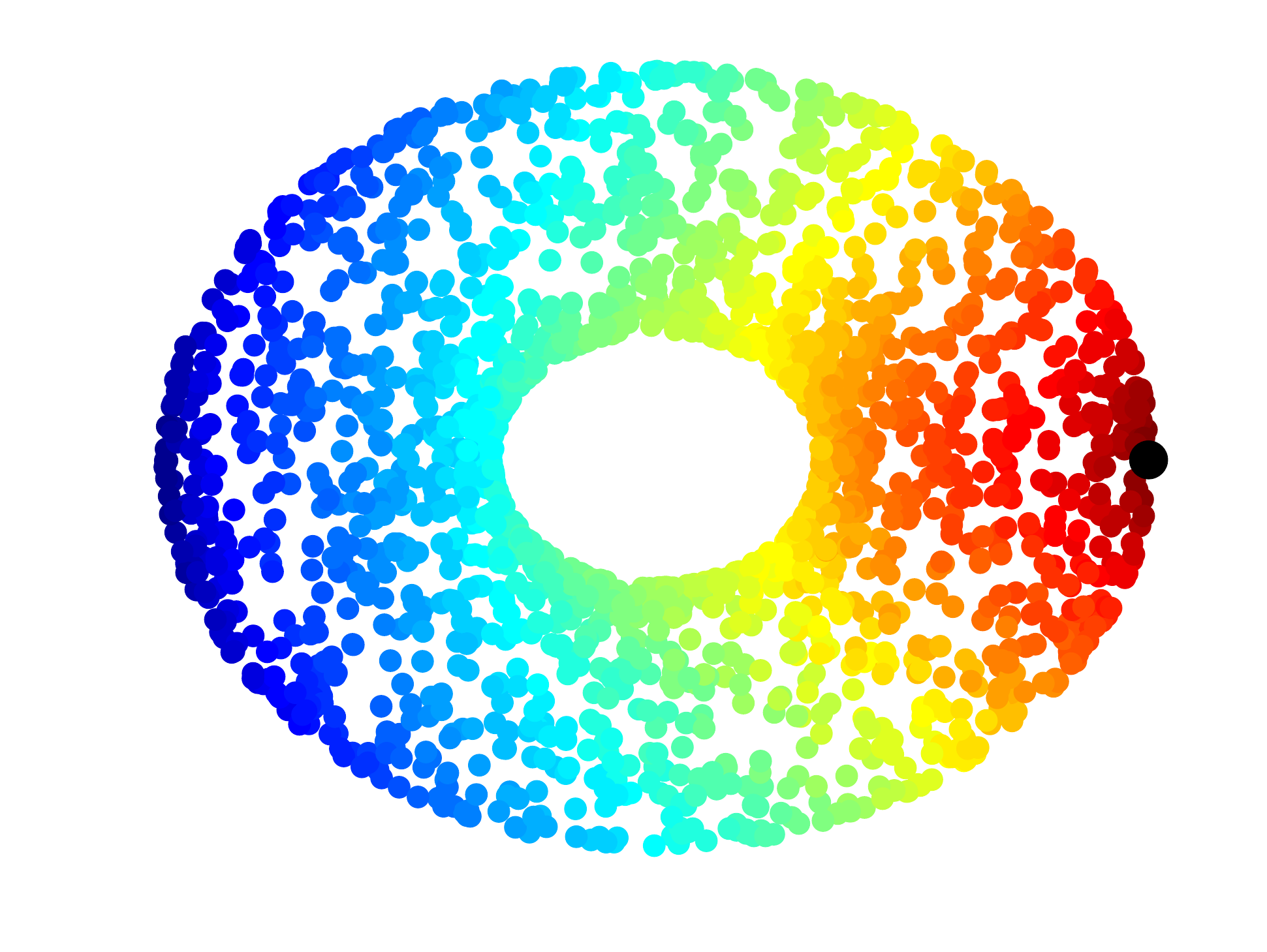}}
\subfigure[PFRank (0.20)]{\includegraphics[width=.23\textwidth]{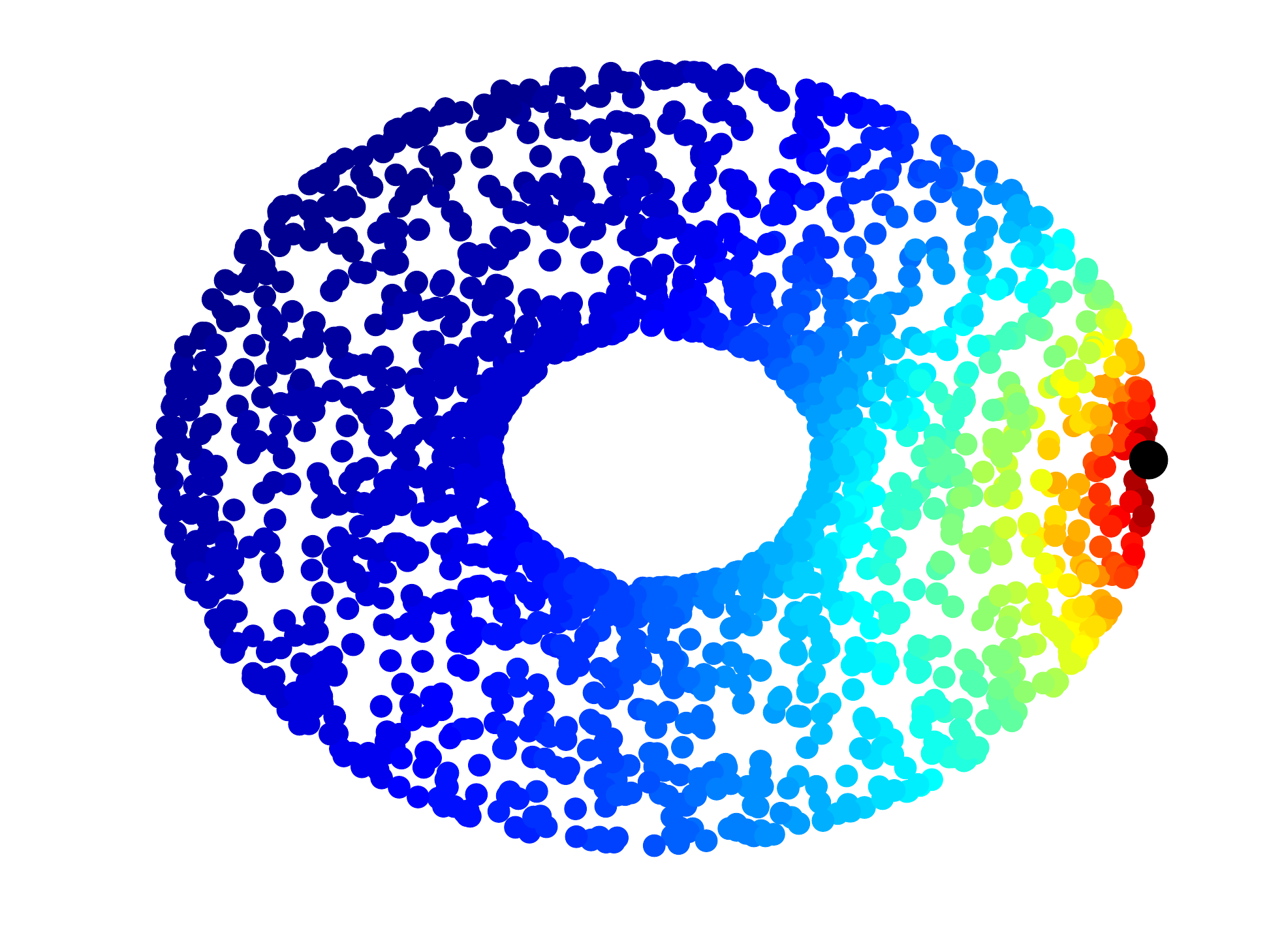}}
\subfigure[MR (0.38)]{\includegraphics[width=.23\textwidth]{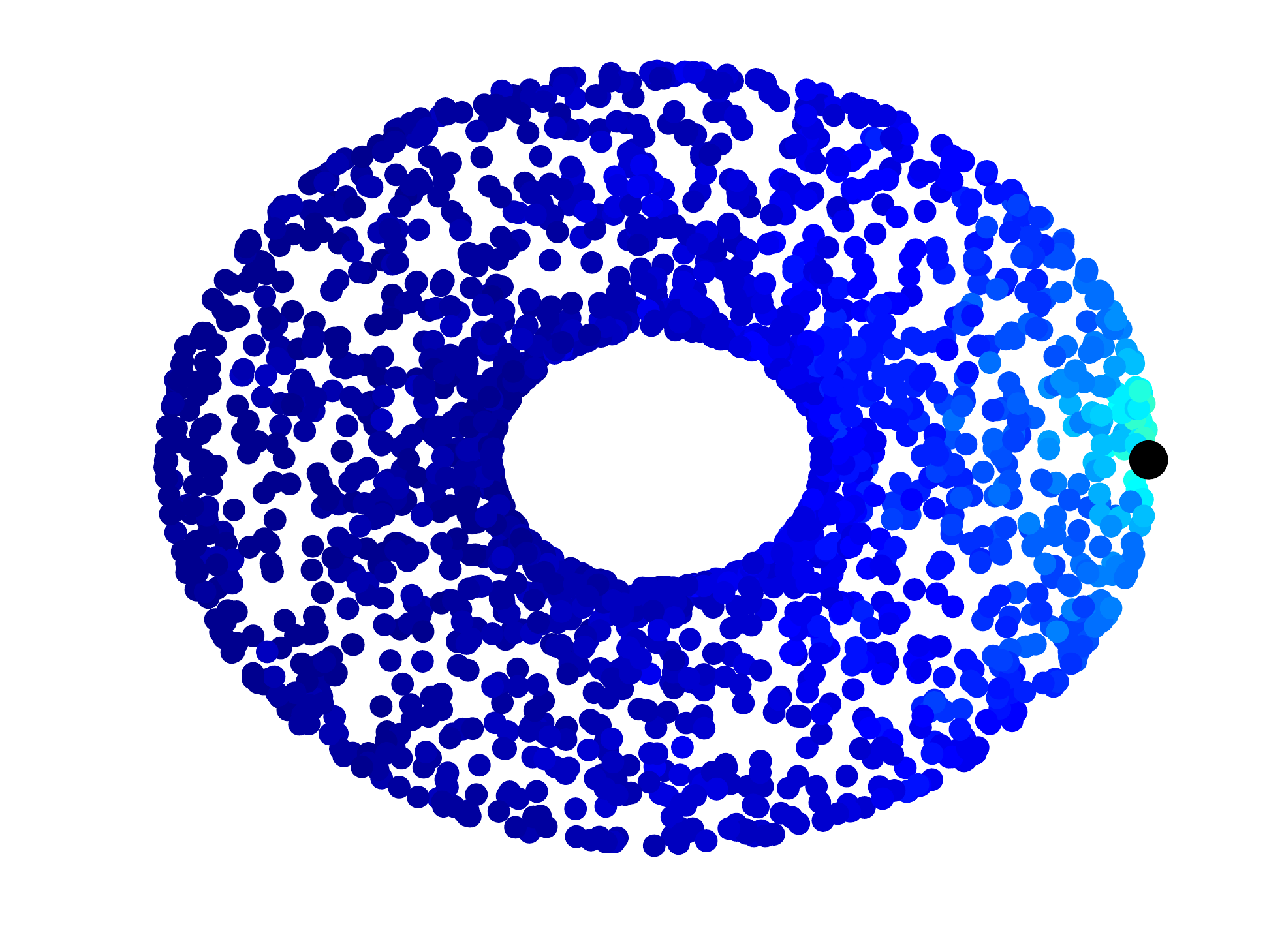}}\\
\subfigure[Vector field by GDL]{\includegraphics[width=.23\textwidth]{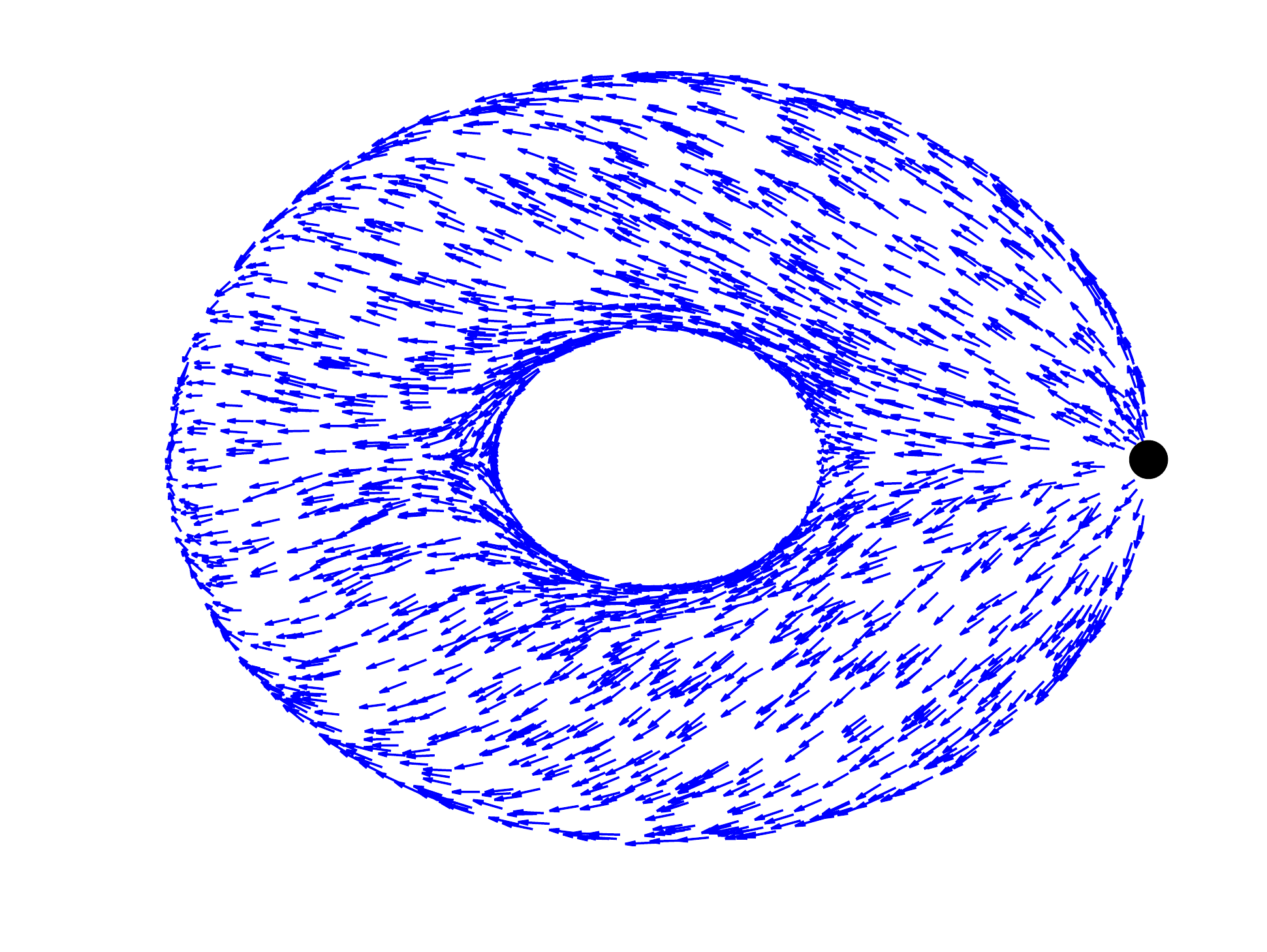}}
\subfigure[HLLE (0.07)]{\includegraphics[width=.23\textwidth]{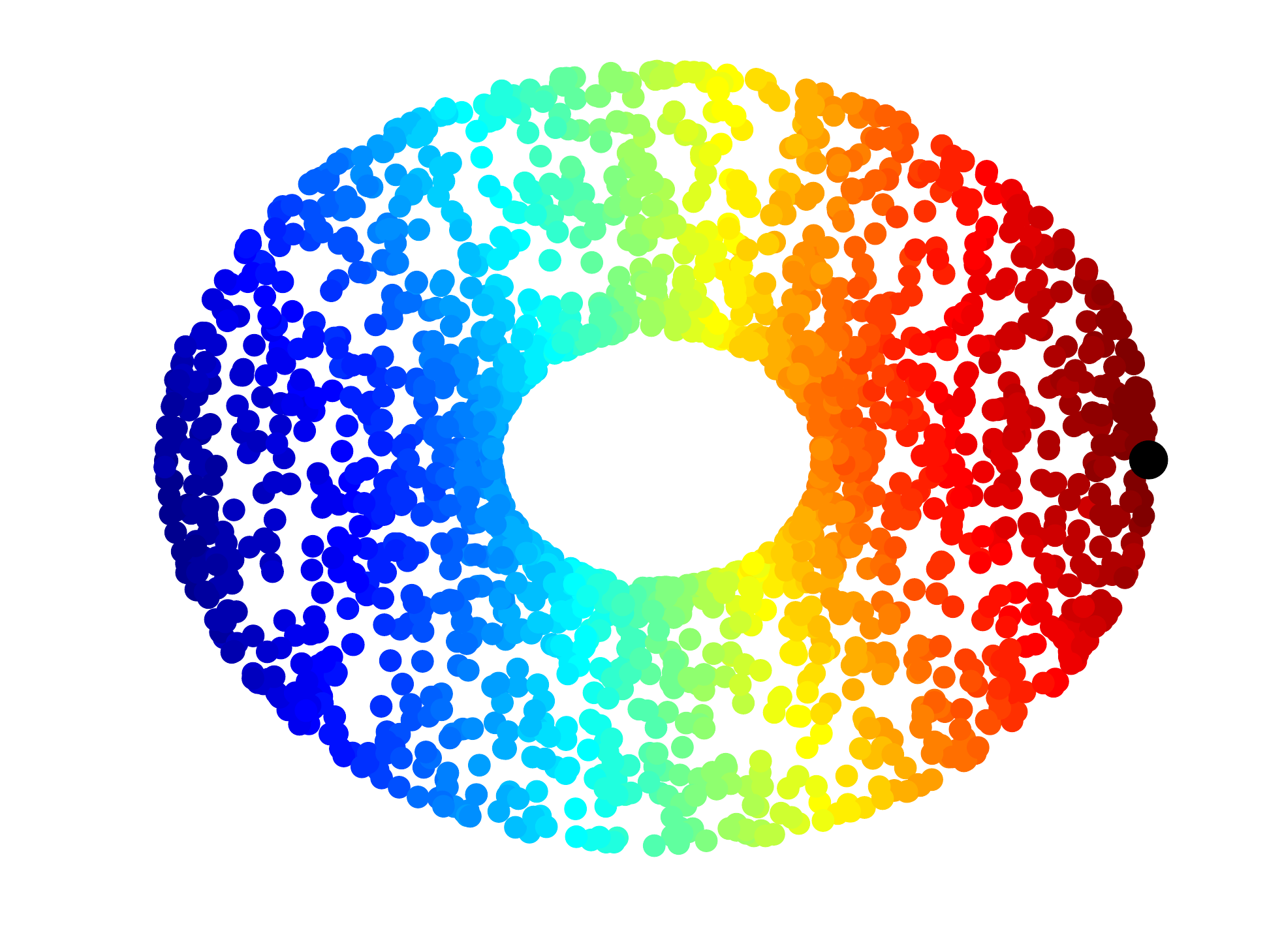}}
\subfigure[LE (0.11)]{\includegraphics[width=.23\textwidth]{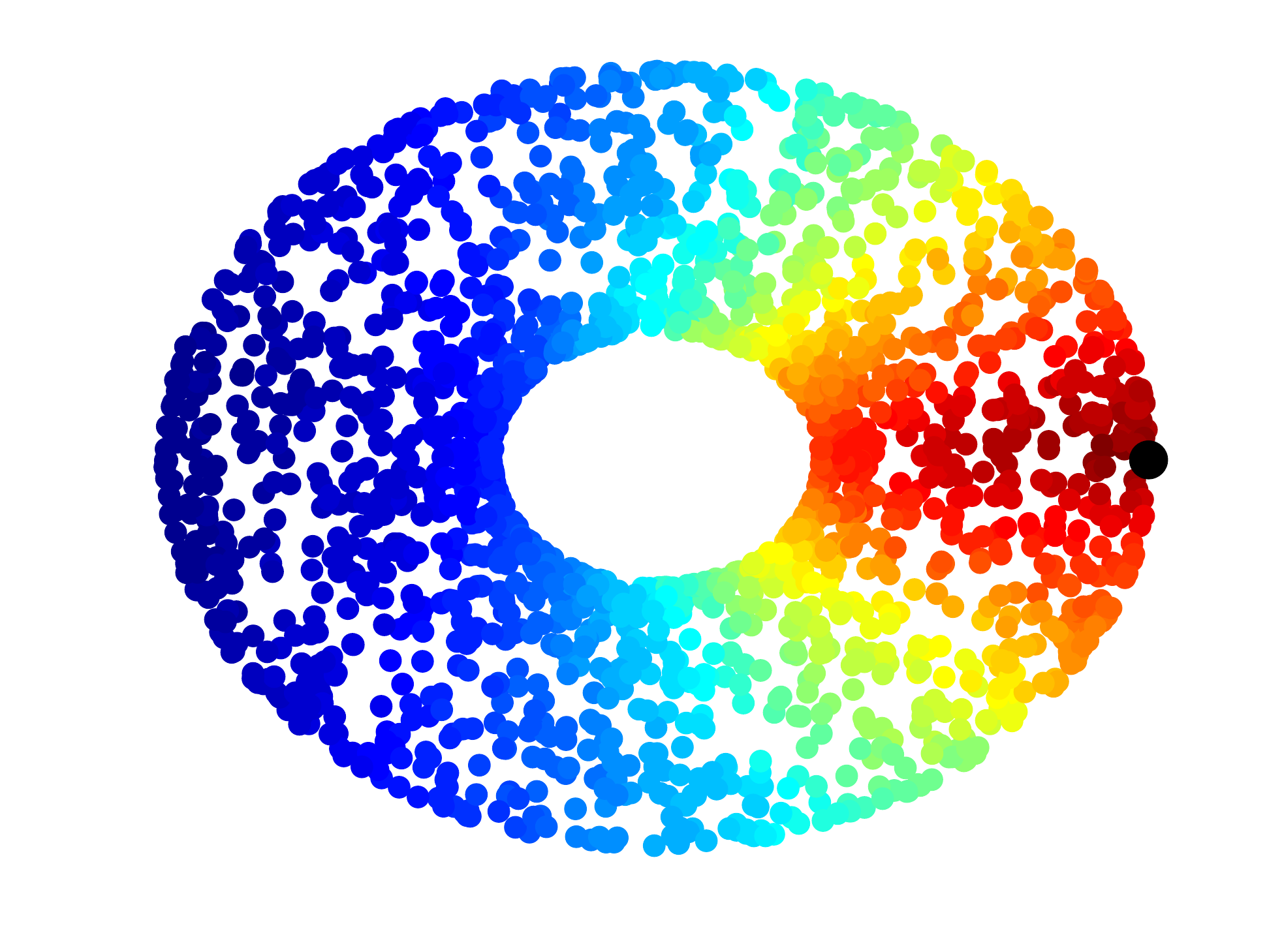}}
\subfigure[MVU (0.05)]{\includegraphics[width=.23\textwidth]{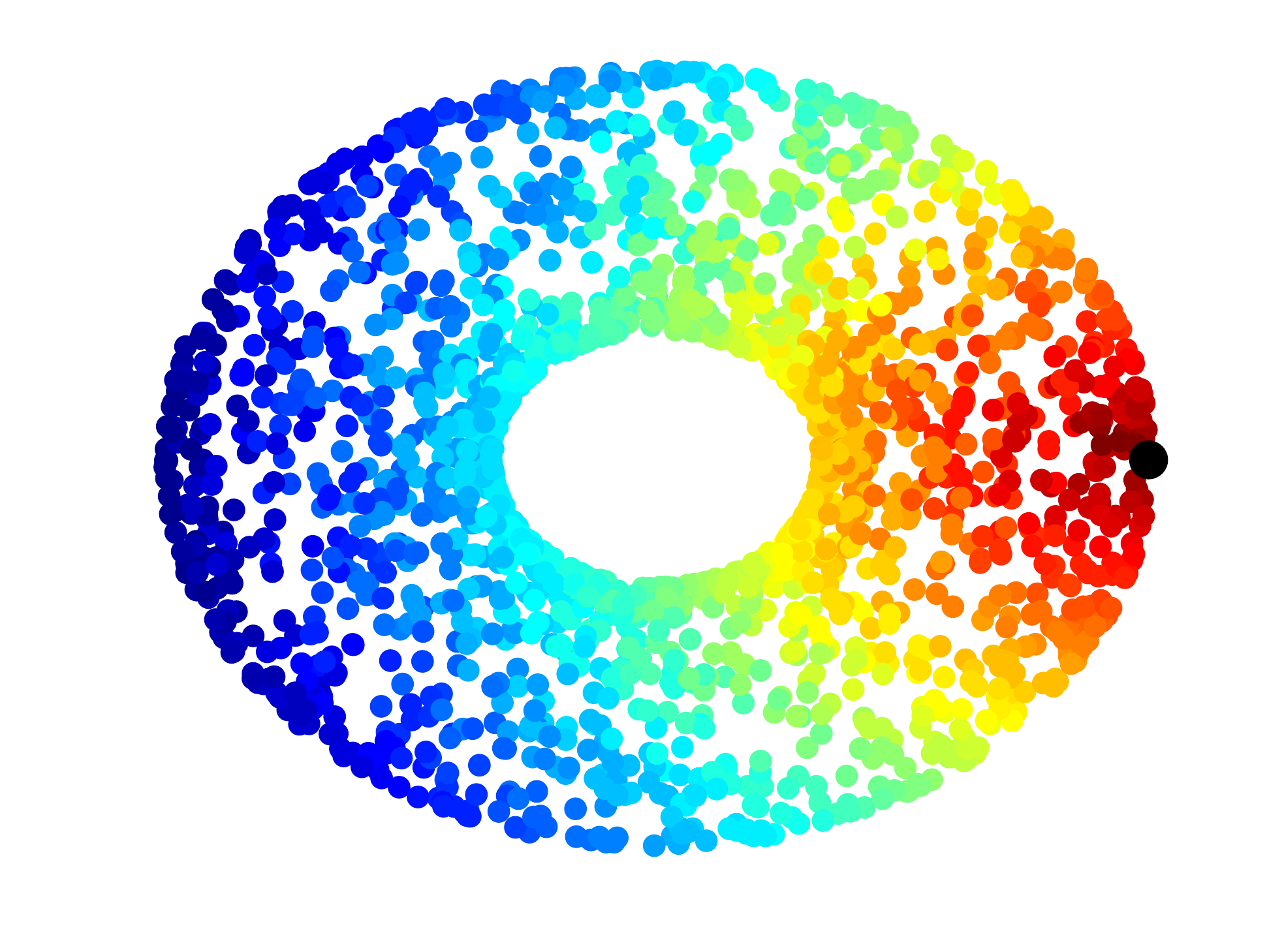}}
\vspace{-0.1in}
\caption{The base point is marked in black. (a) shows the ground truth geodesic distance function, where  (b)-(d) and (f)-(h) visualize the distance functions learned by different algorithms. Different colors indicates different distance values. The number in the brackets measures the difference between the learned distance function and the ground truth.} \label{fig:torus}
\vspace{-0.2in}
\end{center}
\end{figure}

\section{Experiments}
In this section, we empirically evaluate the effectiveness of our proposed Geodesic Distance Learning (GDL) algorithm in comparison with three representative distance metric learning algorithms: Laplacian Eigenmaps (LE, \cite{eigenmap}), Maximum Variance Unfolding (MVU,~\cite{MVU}) and Hessian Eigenmaps (HLLE,~\cite{HLLE}) as well as two state-of-art ranking algorithms: Manifold Ranking (MR,~\cite{Ranking-on-Data-Manifolds}) and Parallel Field Rank (PFRank,~\cite{Ji:2012:PFRank}). As LE, MVU and HLLE cannot directly obtain the distance function, we compute the embedding first and then compute the Euclidean distance between data points in the embedded Euclidean space.

We empirically set $t = 1$ for GDL in all experiments as GDL performs very stable when $t$ varies. The dimension of the manifold $d$ is set to 2 in the synthetic example. For real data, we perform cross-validation to choose $d$. Specifically, $d=9$ for the CMU PIE data set and $d=2$ for the Corel data set. We use the same nearest neighbor graph for all six algorithms. The number of nearest neighbors is set to 16 on both synthetic and real data sets and the weight is the simple $0-1$ weight.

\subsection{Geodesic Distance Learning}

A simple synthetic example is given in Fig.~\ref{fig:torus}. We randomly sample 2000 data points from a torus. It is a 2-dimensional manifold in the 3-dimensional Euclidean space. The base point is marked by the black dot on the right side of the torus. Figs.~\ref{fig:torus}(a) shows the ground truth geodesic distance function which is computed by shortest path distance. Fig.~\ref{fig:torus}(b)-(d) and (f)-(h) visualize the distances functions learned by different algorithms respectively. To better evaluate the results, we compute the error by using the equation $\frac{1}{n} \sum_{i=1}^n | f(x_i) - d(x_q, x_i) |$, where $f(x_i)$ represents the learned distance and $\{ d(x_q, x_i)\}$ represents the ground truth distance. To remove the effect of scale, $\{ f(x_i) \}$ and $\{ d(x_q, x_i)\}$ are rescaled to the range $[0,1]$. As can be seen from Fig.~\ref{fig:torus}, GDL better preserves the distance metric on the torus. Although MVU is comparable to GDL in this example, GDL is approximately thirty times faster than MVU. It might be worth noting that both MR and PFRank achieve poor performance since they are deigned to preserve the ranking order but not the distance.

\subsection{Image Retrieval}
In this section, we apply our GDL algorithm to the image retrieval problem in real world image databases. Two real world data sets are used in our experiments. The first one is from the CMU PIE face database~\cite{PIEfaceDatabase}, which contains $32\times 32$ cropped face images of 68 persons. We choose the frontal pose (C27) with varying lighting conditions, which leaves us 42 images per person. The second data set contains 5,000 images of 50 semantic categories, from the Corel database. Each image is extracted to be a 297-dimensional feature vector. Both of the two image data sets we use have category labels. For each data set, we randomly choose 10 images from each category as queries, and average the retrieval performance over all the queries.

We use precision, recall, and Mean Average Precision (MAP,~\cite{Introduction-IR}) to evaluate the retrieval results of different algorithms. Precision is defined as the number of relevant presented images divided by the number of presented images. Recall is defined as the number of relevant presented images divided by the total number of relevant images in our database. Given a query, let $r_i$ be the relevance score of the image ranked at position $i$, where $r_i = 1$ if the image is relevant to the query and $r_i = 0$ otherwise. Then we can compute the Average Precision (AP):
\begin{equation}
\hbox{AP} = \frac{\sum_{i}r_i\times \hbox{Precision@} i}{\hbox{\# of
relevant images}}.
\end{equation}
MAP is the average of AP over all the queries.

\begin{figure}[t]\small
\begin{center}
\subfigure[Precision-scope curves on PIE]{
\label{fig:pspie}
\includegraphics[width=0.45\linewidth]{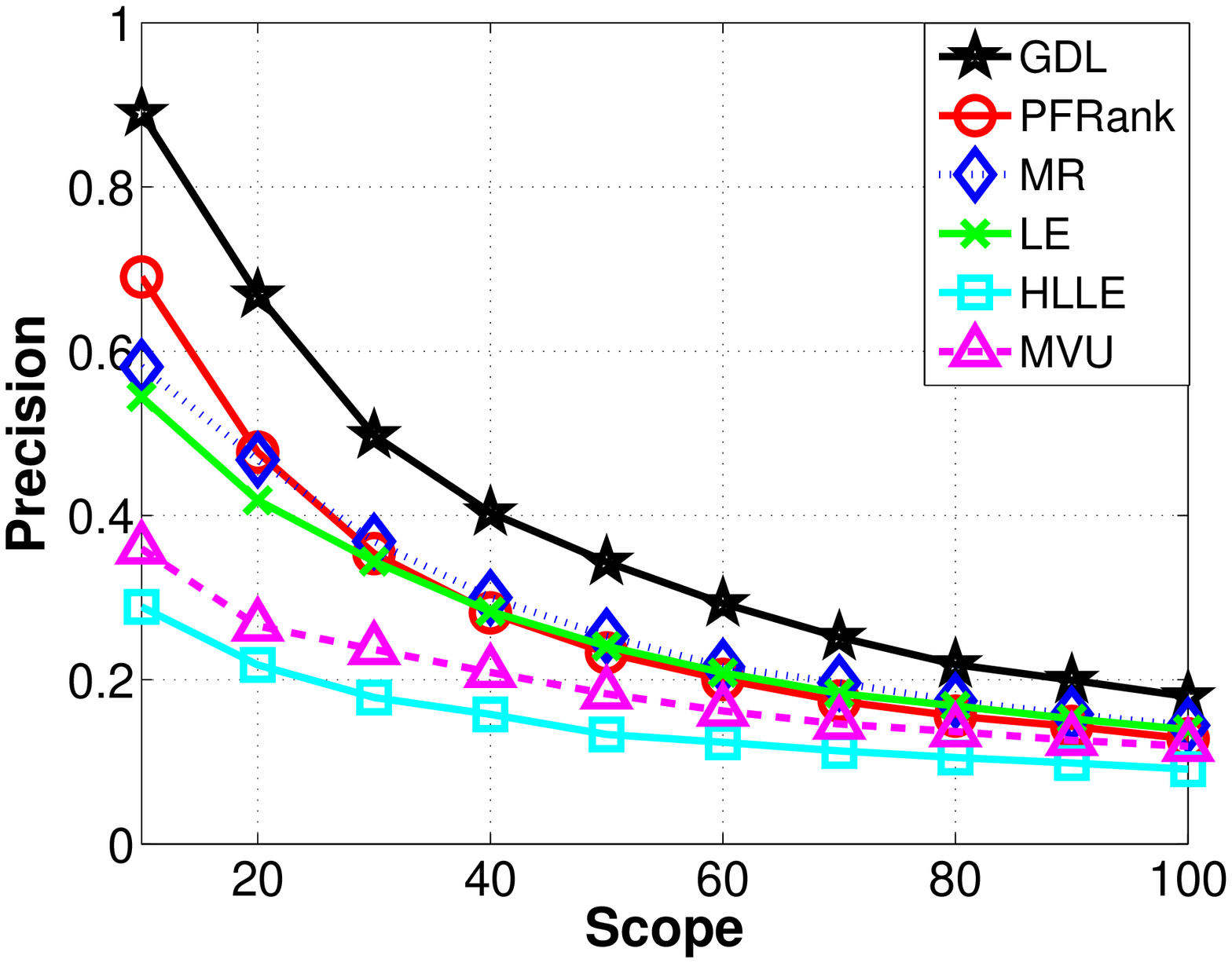}}
\subfigure[Precision-scope curves on Corel]{
\label{fig:pscorel}
\includegraphics[width=0.45\linewidth]{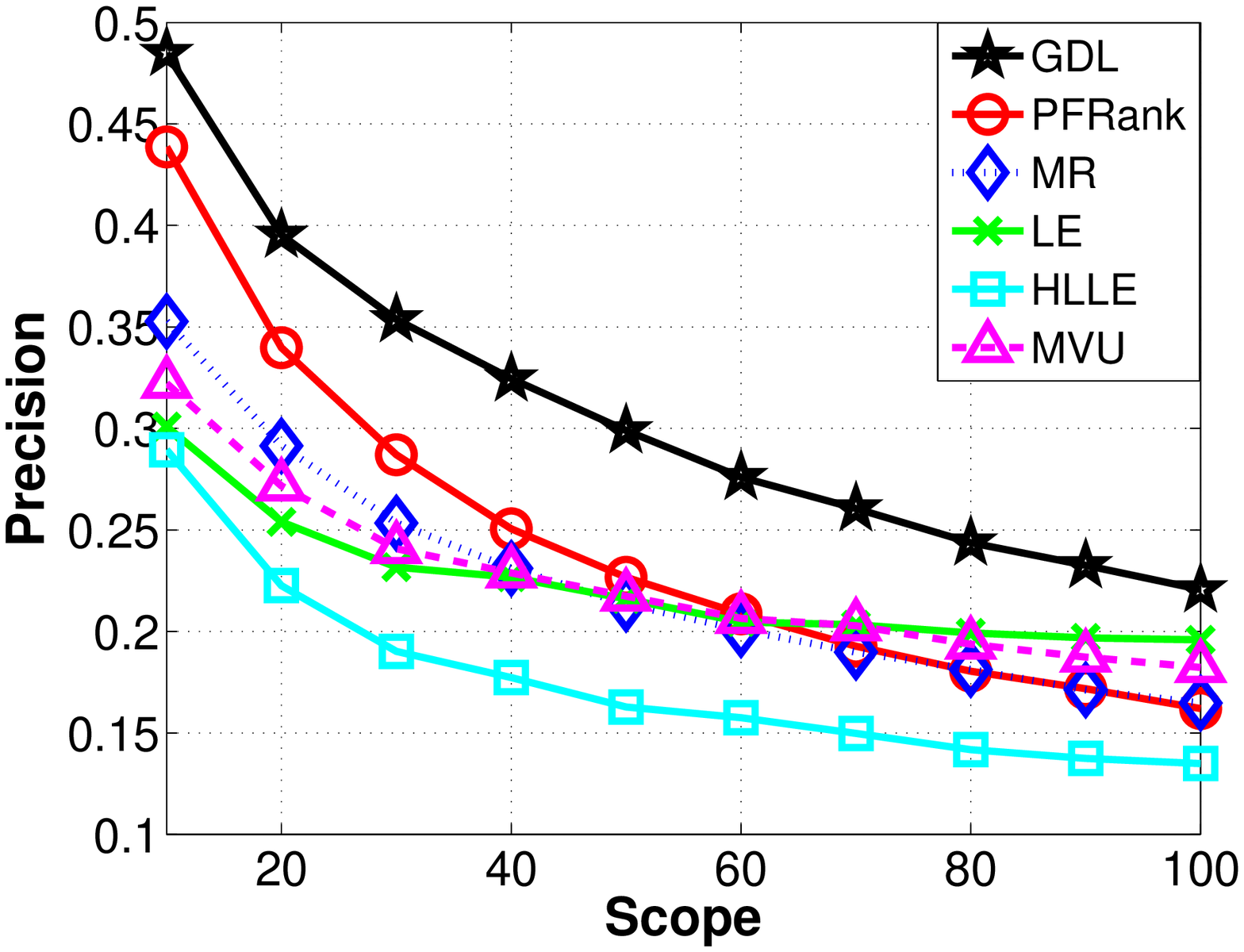}}
\end{center}
\vspace{-0.2in}
\caption{Precision-scope curves.}
\label{fig:performance}
\vspace{-0.05in}
\end{figure}

\begin{table}
\caption{Recall and MAP on the PIE data set.}
\centering
\vskip 0.15in
\begin{tabular}{|c||c|c|c||c|}
\hline Recall & @10 & @20 & @50 & MAP\\
\hline\hline
GDL & \textbf{0.457} & \textbf{0.618} & \textbf{0.783} & \textbf{0.698} \\
\hline
PFRank & 0.443 & 0.585 & 0.713 & 0.596\\
\hline
MR & 0.323 & 0.524 & 0.698 & 0.507 \\
\hline
LE & 0.301 & 0.452 & 0.643 & 0.479 \\
\hline
HLLE & 0.162 &0.234 & 0.357 & 0.245 \\
\hline
MVU & 0.228 & 0.333 & 0.565 & 0.338 \\
\hline
\end{tabular}\label{tab:otherpie}
\caption{Recall and MAP on the Corel data set.}
\centering
\vskip 0.15in
\begin{tabular}{|c||c|c|c||c|}
\hline Recall & @10 & @20 & @50 & MAP\\
\hline\hline
GDL & \textbf{0.134} & \textbf{0.195} & \textbf{0.330} & \textbf{0.340} \\
\hline
PFRank & 0.124 & 0.173 & 0.268 & 0.266\\
\hline
MR & 0.098 & 0.148 & 0.250 & 0.263\\
\hline
LE & 0.092 & 0.127 & 0.233 & 0.268 \\
\hline
HLLE & 0.099 &0.134 & 0.213 & 0.220 \\
\hline
MVU & 0.099 & 0.137 & 0.239 & 0.272 \\
\hline
\end{tabular}\label{tab:othercorel}
\vspace{-0.1in}
\end{table}

Fig.~\ref{fig:pspie} and Fig.~\ref{fig:pscorel} show the average precision-scope curves of various methods on the two data sets, respectively. The scope means the number of top-ranked images returned to the user. The precision-scope curves describe the precision with various scopes, and therefore provide an overall performance evaluation of the algorithms. As can be seen from Fig.~\ref{fig:pspie} and Fig.~\ref{fig:pscorel}, our proposed GDL algorithm outperforms all the other algorithms. We also present the recall and MAP scores of different algorithms on the two data sets in Table~\ref{tab:otherpie} and Table~\ref{tab:othercorel}, respectively. MAP provides a single figure measure of quality across all the recall levels. Our GDL achieves the highest MAP, indicating reliable performance over the entire ranking list. We also performed comprehensive $t$-test with $99\%$ confidence level. The improvements of GDL compared to PFRank and other algorithms are significant with most of the $p$-values less than $10^{-3}$, including those in Fig.~\ref{fig:pspie}, Fig.~\ref{fig:pscorel}, Table~\ref{tab:otherpie} and Table~\ref{tab:othercorel}. These results indicate that learning the distance function directly on the manifold might be better than learning the distance function after embedding. 


\section{Conclusion}
In this paper, we study the geodesic distance from the vector field perspective. We provide theoretical analysis to precisely characterize the geodesic distance function and propose a novel heat flow on vector fields approach to learn it. Our experimental results on synthetic and real data demonstrate the effectiveness of the proposed method. The future work includes developing the machine learning theory and designing efficient algorithms using heat flow on vector fields as well as other general partial differential equations.

\section{Appendix. Justification}\label{sec:justification}

We first show that solving Eq.~\eqref{eq:vf-continuous-obj} is equivalent to solving the heat equation on vector fields. According to the Bochner technique~\cite{RiemannianGeometry}, with appropriate boundary conditions we have
$
\int_{\mathcal{M}} \| \nabla V \|_{\mathrm{HS}}^2 dx = \int_{\mathcal{M}} g( V, \nabla^*\nabla V ) dx,
$
where $\nabla^*\nabla$ is the \emph{connection Laplacian} operator. Define the inner product $(\cdot, \cdot)$ on the space of vector fields as
$
(X, Y) = \int_{\mathcal{M}} g(X, Y) dx.
$
Then we can rewrite $E(V)$ as
$
E(V) = (V-V^0, V-V^0) + t(V, \nabla^*\nabla V).
$
The necessary condition of $E(V)$ to have an extremum at $V$ is that the functional derivative ${\delta E(V)}/{\delta V} = 0$~\cite{Abraham:1988:manifolds}. Using the calculus rules of the functional derivative and the fact $\nabla^*\nabla$ is a self-adjoint operator, we have ${\delta E(V)}/{\delta V} = 2 V - 2 V^0 + 2 t \nabla^*\nabla V$. A detailed derivation can be found in the appendix. Since $\nabla^*\nabla$ is also a positive semi-definite operator, the optimal $V$ is then given by:
\begin{equation}\label{eq:vf-continuous-solution}
 V = (I +  t \nabla^*\nabla)^{-1}V^0,
\end{equation}
where $I$ is the identity operator on vector fields. Let $X(t)$ be a vector field valued function. That is, for each $t$, $X(t)$ is a vector field on the manifold. Given an initial vector field $X(t)|_{t=0} = X_0$, the heat equation on vector fields~\cite{BGV:2004} is given by
$
\frac{\partial X(t)}{\partial t} + \nabla^*\nabla X(t) = 0.
$
When $t$ is small, we can discrete it as follows:
$
\frac{X(t) - X_0}{t} + \nabla^*\nabla X(t) = 0.
$
Then $X(t)$ can be solved as
\begin{equation}\label{eq:discrete-heat-equation}
X(t) = (I + t \nabla^*\nabla)^{-1} X_0.
\end{equation}
If we set $X_0 = V^0$, then Eq.~\eqref{eq:discrete-heat-equation} is exactly the same as Eq.~\eqref{eq:vf-continuous-solution}. Therefore when $t$ is small, solving Eq~\eqref{eq:vf-continuous-obj} is essentially solving the heat equation on vector fields.


Next we analyze the asymptotic behavior of $X(t)$ and show that the heat equation transfers the initial vector field primarily along geodesics. Let $x, y \in \cM$, then $X(t)$ can be obtained via the heat kernel as $X(t)(x) = \int_{\mathcal{M}} k(t, x, y) X_0(y)dy$, where $k(t, x, y)$ is the heat kernel for the connection Laplacian. It is well known for small $t$, we have the asymptotic expansion of the heat kernel~\cite{BGV:2004}:
\begin{equation}
k(t, x, y) \approx (\frac{1}{4\pi t})^{\frac{d}{2}} e^{-d(x, y)^2 / {4t}} \tau(x, y) ,
\end{equation}
where $d(\cdot, \cdot)$ is the distance function, $\tau: T_y{\mathcal{M}} \rightarrow T_x{\mathcal{M}}$ is the parallel transport along the geodesic connecting $x$ and $y$.

\begin{figure}[ht]
\vskip 0.2in
\begin{center}
\centerline{\includegraphics[width=0.7\columnwidth]{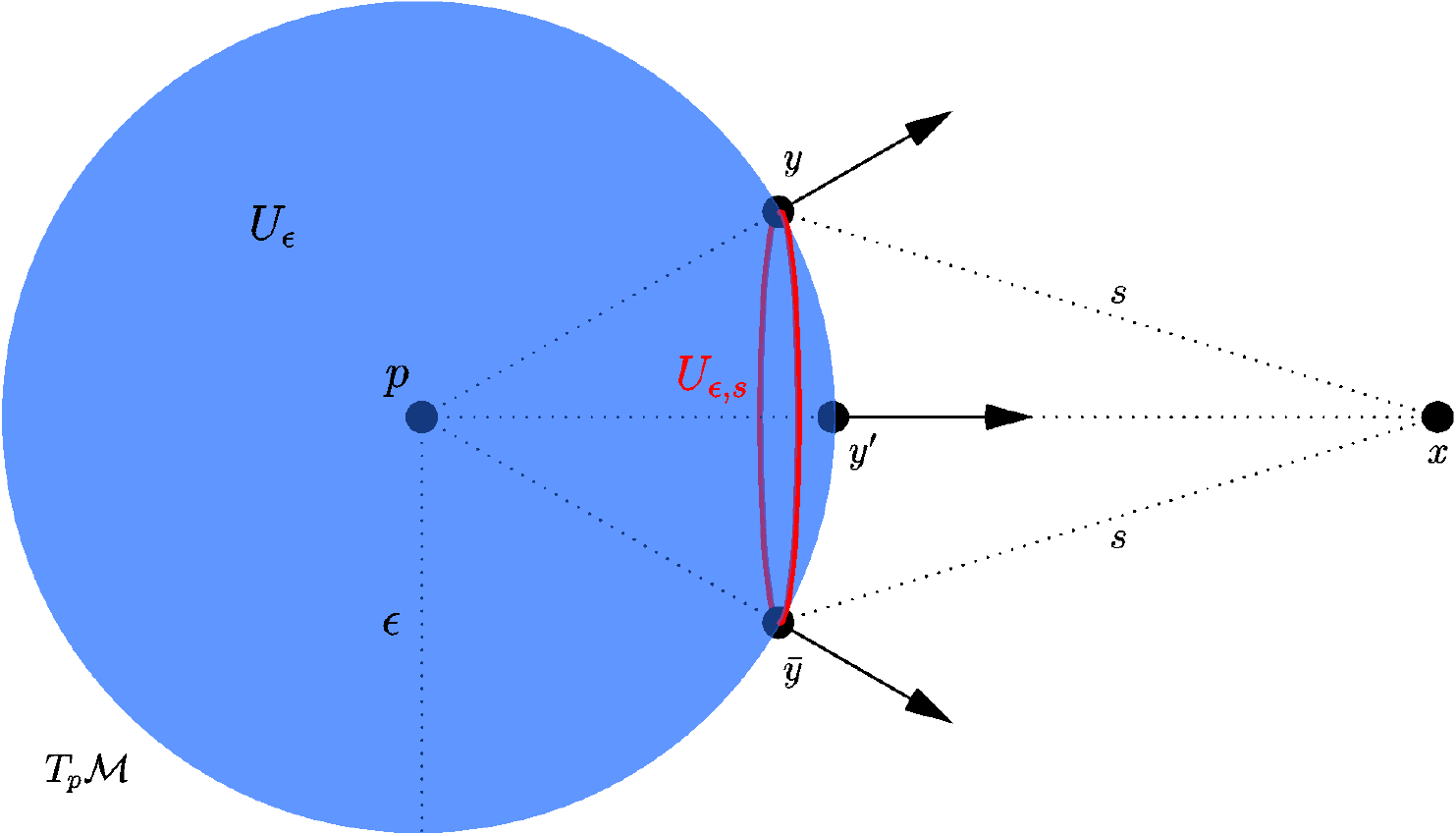}}
\caption{Illustration of heat flow.}
\label{fig:heat-flow-geodesic}
\end{center}
\vskip -0.2in
\end{figure}

Now we consider $X_0 = V^0 $. By construction $X_0(y) = 0$ if $y\notin U_{\epsilon}$. Then the vector $X(t)(x)=\int_{U_{\epsilon}} e^{-d(x, y)^2 / {4t}} \tau(x, y) X_0(y) dy$ up to a scale. To analyze what $X(t)(x)$ is, we first map the manifold $\cM$ to the tangent space $T_p\cM$ by using $\exp_{p}^{-1}$. Then ${U_{\epsilon}}$ becomes a ball in $T_p\cM$; please see Fig.~\ref{fig:heat-flow-geodesic}. In the following we will still use $x$ and $U_{\epsilon}$ to represent $\exp_p^{-1}(x)$ and $\exp_{p}^{-1}(U_{\epsilon})$ for simplicity of notation. Given any point $x \in T_{p}\cM$, we can decompose the ball $U_{\epsilon}$ as $U_{\epsilon} = \cup_{\epsilon', s} U_{\epsilon', s}$ where $U_{\epsilon', s} := \{ y|d(p, y) = \epsilon', d(x, y) = s \}, \epsilon'\leq \epsilon$ and $ 0 \leq s \leq \infty$. Then each section $U_{\epsilon', s}$ is a sphere centered at some point lying on the line segment connecting $p$ and $x$. Therefore $U_{\epsilon', s}$ is symmetric with respect to the vector $x-p$. For any $y \in U_{\epsilon', s}$, there is a unique reflection point $\bar{y}$ such that $\tau(x, y) X_0(y) + \tau(x, \bar{y}) X_0(\bar{y}) $ is parallel to $\tau(x, y')X_0(y')$ where $y' = \arg \min_{y\in U_{\epsilon}} d(x, y)$. Note that the weight $e^{-d(x, y)^2 / {4t}}$ is the same on the section $U_{\epsilon', s}$. We conclude that $ \int_{U_{\epsilon', s}} e^{-d(x, y)^2 / {4t}} \tau(x, y) X_0(y) dy$ is parallel to $\tau(x, y')X_0(y')$. Since $\int_{U_{\epsilon}} = \int_{\epsilon'}\int_s \int_{U_{\epsilon', s}} $, $ X_0(x) \approx \int_{U_{\epsilon}} e^{-d(x, y)^2 / {4t}} \tau(x, y) X_0(y) dy$ is parallel to $\tau(x, y')X_0(y')$. In other words, the vector field flows primarily along geodesics. Therefore given an initial distance vector field around the base point, solving the heat equation will get a vector field which is approximately parallel to the gradient field of the distance function at each point. We can further normalize the vector field at each point to obtain the gradient field of the distance function. From this heat equation point of view, it also provides guidance on the algorithm setting. Specifically, we should set the initial vector field uniformly around the base point and set a small $t$.

\section{Appendix. Backgrounds on Riemannian Geometry}
Let $(\mathcal{M},g)$ be a $d$-dimensional Riemannian manifold, where $g$ is a Riemannian \emph{metric tensor} on $\mathcal{M}$. The goal of \emph{distance metric learning} on the manifold is to find a desired distance function $d(x, y)$ such that it provides a natural measure for the \emph{similarity} between two data points $x$ and $y$ on the manifold. A distance function $d(\cdot, \cdot)$ defined by its Riemannian metric $g$ is often called an intrinsic distance function. In this paper, we study a fundamental intrinsic distance function - the geodesic distance function. Similar to many differential geometry textbooks~(e.g., \cite{Jost08Geometry, RiemannianGeometry}), we simply call it the distance function.


\subsection{Tangent Spaces and Vector Fields}
As the manifold is locally a Euclidean space, the key tool for studying the manifold will be the idea of \emph{linear approximation}. The fundamental linear structure of the manifold is the tangent space.
\begin{Def}[Tangent space,~\cite{LEESmoothManifold}]
Let $\cM$ be a smooth manifold and let $p$ be a point on $\cM$. A linear map $X: C^{\infty}(\cM) \rightarrow \mathbb{R}$ is called a derivation at $p$ if it satisfies
\[
X(fg) = f(p)Xg + g(p)Xf
\]
for all smooth functions $f, g\in C^{\infty}(\cM)$. The set of all derivations of $C^{\infty}(\cM)$ at $p$ is a vector space called the \emph{tangent space} to $\cM$ at $p$, and is denoted by $T_p\cM$. An element of $T_p\cM$ is called a \emph{tangent vector} at $p$.
\label{def:ts}
\end{Def}
The definition of the tangent space is totally abstract. We first take an example in Euclidean space to show that why a tangent vector is a derivation. Let $v$ denote a geometric tangent vector in $\mathbb{R}^m$. Define a map $D_v|_a : C^{\infty}(\mathbb{R}^m) \rightarrow \mathbb{R}$, which takes the directional derivative in the direction $v$ at $a$:
\[
D_v|_a f = D_v f(a) := \frac{d}{dt}|_{t=0}f( a + tv ).
\]
Clearly this operation is linear and it satisfies the derivation rule. Therefore we might write the directional derivative of $f$ in the direction of $Y$ as $Yf = Y(f) = D_Y f = \nabla_Y f$, where $\nabla$ denotes the covariant derivative on the manifold. Next we show what a tangent space is on the manifold by using local coordinates. Let $\{x^i| i=1,\ldots, d\}$ denote a local coordinate chart around $p$. Then it can be easily verified by definition that $\partial_i|_p:=\frac{\partial}{\partial x_i}|_p$ is a tangent vector at $p$. Moreover, these coordinate vectors ${\partial_1|_p, \ldots, \partial_d|_p}$ form a basis for $T_p\cM$~\cite{LEESmoothManifold}. Therefore, the dimension of the tangent space is exactly the same as the dimension of the manifold. For example, consider a two dimensional sphere embedded in $\mathbb{R}^3$; given any point of the sphere, the tangent space of the sphere is just a two dimensional plane.

For any smooth manifold $\cM$, we define the \emph{tangent bundle} of $\cM$, denoted by $T\cM$, to be the disjoint union of the tangent spaces at all points of $\cM$: $T\cM = \cup_{p\in \cM}T_p\cM.$ Now we define the vector field.
\begin{Def}[Vector field,~\cite{LEESmoothManifold}]\label{def:vector-fields}
A vector field is a continuous map $X:\mathcal{M} \rightarrow T\mathcal{M}$, usually written as $p \mapsto X_p$, with the property that for each $p\in \mathcal{M}$, $X_p$ is an element of $T_p\mathcal{M}$.
\end{Def}
Intuitively, a vector field is nothing but a collection of tangent vectors with the continuous constraint. Since at each point, a tangent vector is a derivation. A vector field can be viewed as a \emph{directional derivative} on the whole manifold. It might be worth noting that each vector $X_p$ of a vector field $X$ must be in the corresponding tangent space $T_p\mathcal{M}$. Let $X$ be a vector field on the manifold. We can represent the vector field locally using the coordinate basis as $X = \sum_{i=1}^d a^i\partial_i$, where each $a^i$ is a function which is often called the coefficient of $X$. For the sake of convenience, we will use the Einstein summation convention: when an index variable appears twice in a single term, it implies summation of that term over all the values of the index, i.e., we might simply write $a^i\partial_i$ instead of $\sum_{i=1}^d a^i\partial_i$.

\subsection{Distance Functions}\label{sec:distance-function}
Next, we show how to assign a metric structure on the manifold. For each point $p$ on the manifold, a Riemannian metric tensor $g$ at $p$ is a Euclidean inner product $g_p$ on each of the tangent space $T_{p}\mathcal{M}$ of $\mathcal{M}$. In addition we assume that $g_p$ varies smoothly~\cite{RiemannianGeometry}. This means that for any two smooth vector fields $X, Y$ the inner product $g_p(X_p, Y_p)$ should be a smooth function of $p$, where $X_p$ and $Y_p$ denote the tangent vectors of $X$ and $Y$ at $p$. The subscript $p$ will be omitted when it is clear from the context. Thus we might write $g(X,Y)$ or $g_p(X, Y)$ with the understanding that this is to be evaluated at each $p$ where $X$ and $Y$ are defined. We define the norm of a tangent vector $v\in T_p\mathcal{M}$ as $\|v\| = \sqrt{g_p(v, v)}$. Once we have defined a metric tensor on the manifold, we can define distance on the manifold. Let $[a,b]$ be a closed interval in $\mathbb{R}$, and $\gamma:[a,b]\rightarrow \mathcal{M}$ be a smooth curve. The \emph{length} of $\gamma$ can then be defined as
$
l(\gamma)= \int_{a}^b \|\frac{d\gamma}{dt}(t)\|dt.
$
The \emph{distance} between two points $p, q$ on the manifold $\mathcal{M}$ can be defined as:
\begin{eqnarray*}
d(p,q) := \inf \{ l(\gamma): \gamma:[a,b]\rightarrow \mathcal{M}~ \mathrm{piecewise~ smooth} \\ \mathrm{curve~ with~} \gamma(a) = p,\gamma(b)=q  \}.
\end{eqnarray*}
We call $d(\cdot,\cdot)$ the \emph{distance function} and it satisfies the usual axioms of a metric, i.e., positivity, symmetry and triangle inequality~\cite{Jost08Geometry}.

Computing the distance function $d(\cdot, \cdot)$ for a given manifold is very difficult. In this paper, we study the distance function $d(p, \cdot)$ when $p$ is given.
\begin{Def}[Distance function based at a point]
Let $\mathcal{M}$ be a Riemannian manifold, and let $p$ be a point in $\mathcal{M}$. A \emph{distance function} on $\mathcal{M}$ based at $p$ is defined as $r_p(x) = d(p, x)$.
\end{Def}
For simplicity, we might write $r(\cdot)$ instead of $r_p(\cdot)$. Recall from the definition of the distance function that $r_p(x)$ measures the minimum path distance between $p$ and $x$. 

\subsection{Covariant Derivative}
A vector field can measure the change of functions on the manifold. Now we consider the question of measuring the change of vector fields. Let $X = a^i\partial_i$ be a vector field in $\mathbb{R}^d$ where $\partial_i$ denotes the standard Cartesian coordinate. Then it is natural to define the \emph{covariant derivative} of $X$ in the direction $Y$ as
\[
\nabla_Y X = (\nabla_Y a^i)\partial_i = Y(a^i)\partial_i.
\]
Therefore we measure the change in $X$ by measuring how the coefficients of $X$ change. However, this definition relies on the fact that the coordinate vector field $\partial_i$ is constant vector field. In other words, $\nabla_{Y}\partial_i = 0$ for any vector field $Y$. For general coordinate vector fields, they are not always constant. Therefore, we should give a coordinate free definition of the covariant derivative.
\begin{Them}[The fundamental theorem of Riemannian geometry,~\cite{RiemannianGeometry}]
The assignment $X\rightarrow \nabla X$ on $(\cM, g)$ is uniquely defined by the following properties:
\begin{enumerate}
  \item $Y\rightarrow \nabla_Y X$ is a $(1, 1)$-tensor:
  \[
  \nabla_{\alpha v + \beta w} X = \alpha \nabla_{v} X + \beta \nabla_{w} X.
  \]
  \item $X \rightarrow \nabla_Y X $ is a derivation:
  \[
\begin{aligned}
  \nabla_Y(X_1 + X_2) &= \nabla_Y X_1 + \nabla_Y X_2,\\
  \nabla_Y(fX) &= (Yf)X + f\nabla_Y X
\end{aligned}
 \]
  for functions $f:\cM \rightarrow \mathbb{R}$.
  \item Covariant differentiation is torsion free:
  \[
  \nabla_X Y - \nabla_Y X = [ X, Y ].
  \]
  \item Covariant differentiation is metric:
  \[
  Zg(X, Y) = g( \nabla_Z X, Y ) + g(X, \nabla_Z Y),
  \]
  where $Z$ is a vector field.
\end{enumerate}
\end{Them}
Here $[ \cdot, \cdot ]$ denotes the Lie derivative on vector fields defined as $[X, Y] = XY - YX$. Any assignment on a manifold that satisfies rules 1-4 is called an \emph{Riemannian connection}. This connection is uniquely determined by these four rules.

Let us see what a connection is in local coordinates. Let $X$ and $Y$ be two vector fields on the manifold, we can represent them by local coordinates as $X = a^i \partial_i$ and $Y = b^j \partial_j$. Now we can compute $\nabla_Y X$ in local coordinates using the four rules as follows:
\begin{equation}\label{eq:covariant-derivative-local-coordinates}
\nabla_Y X = \nabla_{b^i\partial_i} a^j\partial_j
= b^i\nabla_{\partial_i} a^j\partial_j
= b^i{\partial_i}(a^j)\partial_j + b^i a^j \nabla_{\partial_i} \partial_j.
\end{equation}
The second equality holds due to the first rule of the connection and the third equality holds due to the second rule of the connection. Since $\nabla_{\partial_i} \partial_j$ is still a vector field on the manifold, we can further represent it as $\nabla_{\partial_i} \partial_j = \Gamma^k_{ij}\partial_k$, where $\gamma^k_{ij}$ are called Christoffel symbols~\cite{RiemannianGeometry}. The Christoffel symbols can be represented in terms of the metric.

\subsection{Geodesics}
Let $\gamma: [a, b] \rightarrow \mathcal{M}$ be a curve in $\mathcal{M}$. Let $\{ x^i | i=1,\ldots,d\}$ denote a local coordinate chart of the manifold, then $\partial_i := \frac{\partial}{\partial x_i}$ form a basis of the tangent spaces. We can represent $\gamma$ by the coordinates $\{ x^i \}$ as $\gamma(t) = (\gamma^1(t),\ldots,\gamma^d(t))$. The velocity in the $t$ direction is then given by
$
 \gamma'(t) = \frac{d \gamma^i}{d t}\partial_i.
$
A vector field $V$ along $\gamma$ is by definition a function $V: [a, b] \rightarrow T\mathcal{M}$ with $V(t) \in T_{\gamma(t)}\mathcal{M} $ for all $t\in [a, b]$.  Next we show how to define the derivative of the vector field $V$ along the curve $\gamma(t)$:
$
\nabla_{{\gamma'}} V =\frac{d}{dt}V(t) .
$
Since $V(t)$ is a vector field along the curve $\gamma$, we can represent it by the basis of tangent spaces $\partial_i$ as
$
V(t) = \varphi^i(t) \partial_i,
$
where $\varphi^i(t)$ are the coefficients of the vector field $V(t)$. We define
\begin{equation}\label{eq:connection-along-curve}
\frac{d}{dt}V(t) := \frac{d \varphi^i}{d t} \partial_i + {\varphi}^i  \nabla_{\gamma'}\partial_i,
\end{equation}
where $\nabla_{\cdot} \cdot$ is the covariant derivative on the manifold. One can show that this definition is independent of the choice of $\partial_i$.

\begin{figure}\centering
\includegraphics[width=0.6\linewidth]{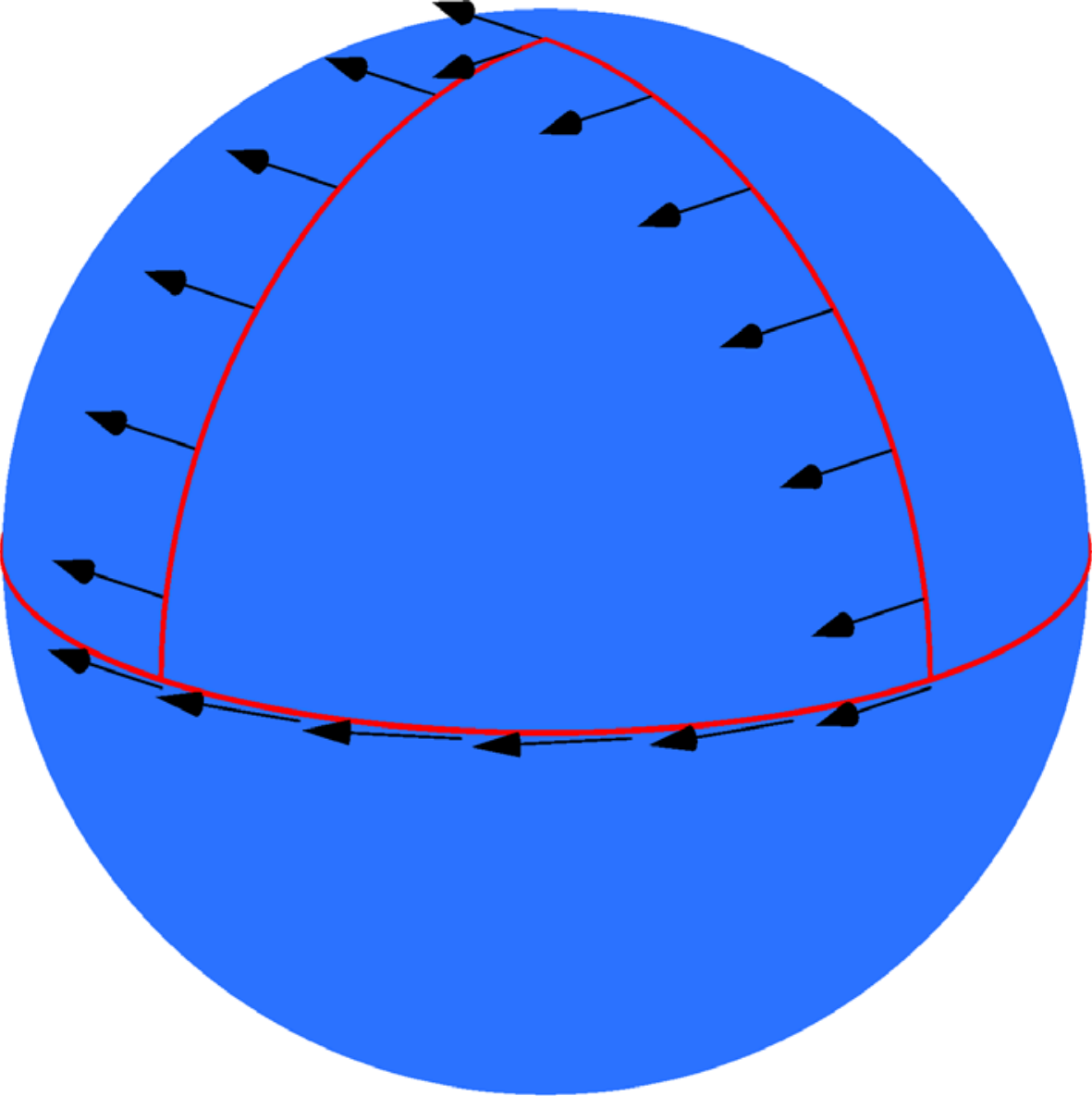}
\hskip 0.1\linewidth
\caption{{An example of the parallel transport}. This figure shows the parallel transport along three different curves. It might be worth noting that there are three parallel fields along three different curves but not one parallel field.}
\label{fig:parallel-transport}
\end{figure}

A vector field $V$ along $\gamma$ is said to be \emph{parallel along} $\gamma$ if $\nabla_{\gamma'} V = 0$. If $V$ and $W$ are parallel fields along $\gamma$, then we have that $g(V, W)$ is constant along $\gamma$. It can be seen from the following: $\frac{d}{d t} g(V, W) =  g(\frac{d}{d t} V, W) + g( V, \frac{d}{d t} W) = g (\nabla_{\gamma'} V, W) + g ( V, \nabla_{\gamma'} W) = 0$. Therefore, parallel fields along a curve do not change their lengths or their angles relative to each other, just as parallel fields in Euclidean space are of constant length and make constant angles.
\begin{Them}[Existence and Uniqueness of Parallel Fields] If $t_0\in [a, b]$ and $v\in T_{\gamma(t_0)}\cM$, then there is a unique parallel field $V(t)$ defined on all of $[a, b]$ with $V(t_0) = v.$
\end{Them}
Therefore, given a curve and an initial vector on it, there exists a parallel field along the curve and such a parallel field is unique. Suppose we are given a vector $v \in T_{\gamma(t_0)}\cM$. We define the \emph{parallel transport} of $v$ along $\gamma$ as the extension of $v$ to a parallel vector field $V$ along $\gamma$. Therefore a parallel transport map $\tau(\gamma(t_1), \gamma(t_2)): T_{\gamma(t_2)}\cM \rightarrow T_{\gamma(t_1)}\cM$ can be defined as follows:
\[
\tau(\gamma(t_1), \gamma(t_2)) V_{\gamma(t_2)} = V(t_1).
\]
We show an example of the parallel transport in Fig.~\ref{fig:parallel-transport}. Three parallel fields along three different curves are presented. 

If we consider $V(t)$ to be ${\gamma'}(t)$ but not a general vector field, then $\nabla_{\gamma'} \gamma' = 0$ means a curve $\gamma$ is parallel along itself. Such a curve is called a \emph{geodesic}. A geodesic can be viewed as a \emph{curved straight line} on the curved manifold.
\begin{Def}[Geodesic,~\cite{RiemannianGeometry}]
Let $\gamma: [a, b]\rightarrow \mathcal{M}, t\mapsto \gamma(t)$ be a smooth curve. $\gamma$ is called a \emph{geodesic} if ${\gamma'}(t)$ is parallel along $\gamma$, i.e., $ \nabla_{\gamma'(t)}\gamma'(t)=0$ for all $t\in [a, b]$.
\end{Def}

Let us see what it is in local coordinates. Recall Eq.~\eqref{eq:connection-along-curve} and note that $\gamma'(t) =  \frac{d \gamma^i}{d t}\partial_i$, we have
\[
\begin{aligned}
 \nabla_{\gamma'(t)}\gamma'(t)
&= \frac{d^2\gamma^i}{dt^2} \partial_i +\frac{d \gamma^i}{d t}  \nabla_{{\gamma'}}\partial_i \\
&=\frac{d^2\gamma^i}{dt^2} \partial_i + \frac{d \gamma^i}{d t} \nabla_{\frac{d\gamma^j}{dt}\partial_j}\partial_i \\
&=\frac{d^2\gamma^i}{dt^2} \partial_i + \frac{d \gamma^i}{d t} \frac{d\gamma^j}{dt}\nabla_{\partial_j}\partial_i \\
&=\frac{d^2\gamma^k}{dt^2} \partial_k + \frac{d \gamma^j}{d t} \frac{d\gamma^i}{dt}\nabla_{\partial_i}\partial_j\\
&=\frac{d^2\gamma^k}{dt^2} \partial_k + \frac{d \gamma^i}{d t} \frac{d\gamma^j}{dt}\Gamma_{ij}^k\partial_k.
\end{aligned}
\]
The first equality holds due to Eq.~\eqref{eq:connection-along-curve} and the third equality holds due to the first rule of the covariant derivative. We can also derive what $\nabla_{\gamma'(t)}\gamma'(t)$ is using Eq.~\eqref{eq:covariant-derivative-local-coordinates} and we will get the same result. As can be seen from the derivation, a geodesic can also be characterized by following equations: $\frac{d^2\gamma^k}{dt^2}  + \frac{d \gamma^i}{d t} \frac{d\gamma^j}{dt}\Gamma_{ij}^k=0$ for $k=1,\ldots, d$. Each equation is a second order nonlinear partial differential equation. If $\gamma$ is a geodesic, the speed of the geodesic $\|\gamma'(t)\|=\sqrt{g(\gamma'(t),\gamma'(t))}$ is constant since parallel fields along a curve have constant speed.

The geodesic and the distance function is related as follows.
\begin{Them}[\cite{RiemannianGeometry}]\label{Them:segments_are_geodesics}
If $\gamma$ is a local minimum for $\inf l(\gamma)$ with fixed end points, then $\gamma$ is a geodesic.
\end{Them}
This theorem tells us that the shortest path on the manifold must be a geodesic. However, the converse is not always true. We call a geodesic a \emph{minimal geodesic} if it has shortest length among all curves with the same end points. Therefore, given any two points $x$ and $y$, the distance function measures the distance of the minimal geodesic connecting $x$ and $y$.

\begin{figure}[h]\centering
\includegraphics[width=0.6\linewidth]{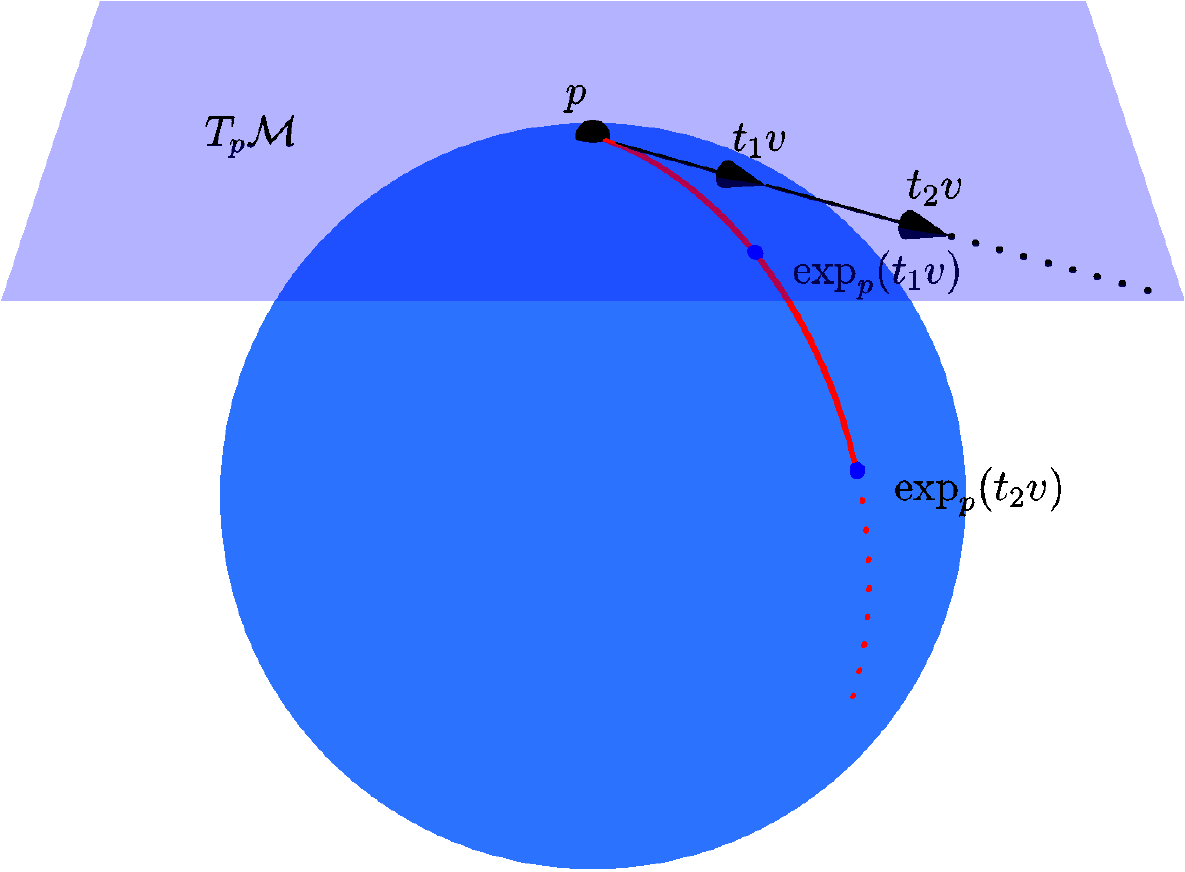}
\hskip 0.1\linewidth
\caption{{An exponential map on a sphere.}}
\label{fig:exponential-map}
\end{figure}

Another important concept for studying geodesics is the \emph{exponential map}.
\begin{Def}[\cite{RiemannianGeometry}]
Let $v \in T_p \mathcal{M}$ be a tangent vector. Then there is a {\em unique} geodesic $\gamma_v$ satisfying $\gamma_v(0)=p$ with the initial tangent vector $\gamma'_v(0) = v$. The corresponding exponential map is defined as
$$
\exp_p(tv)=\gamma_v(t).
$$
\end{Def}
The exponential map $\exp_{p}:T_{p}\mathcal{M}\rightarrow \mathcal{M}$ maps the tangent space $T_{p}\mathcal{M}$ to the manifold $\mathcal{M}$. Given a point $p$, the exponential map is well-defined at $p$ if $\exp_p(tv)$ exists for any $t \geq 0$ and $v\in T_p\cM$. A manifold is called complete or geodesically complete if the exponential map is well-defined at every point of the manifold. When $t$ is small, the exponential map $\exp_p(\cdot v)$ is the minimal geodesic connecting $p$ and $\gamma_v(t)$. When $t$ grows up, this property may not hold. Please see Fig.~\ref{fig:exponential-map} as an example. When $t$ is small, the exponential map $\exp_p(t v)$ is the minimal geodesic connecting $p$ and $\gamma_v(t)$. When $t$ grows up, the exponential map $\exp_p(t v)$ will pass through the opposite point of $p$. Then $\exp_p(t v)$ is no longer a minimal geodesic connecting $p$ and $\gamma_v(t)$. Let $t_0 = \sup \{ t>0 |~\gamma$ is the unique minimal geodesic connecting $p$ and $ \gamma_v(t) \}$. If $t_0<+\infty$, we call $\gamma_v(t_0)$ the cut point of $p$. All cut points of $p$ constitute the \emph{cut locus} of $p$ and we denote it as $\mathrm{Cut}(p)$. Given a point $p$, there is at most one cut point for each unit vector $v\in T_p\cM$. Since the space of unit vector in $T_p\cM$ is a $(d-1)$-dimensional unit sphere, the dimension of the cut locus is at most $d-1$. Therefore, the measure of the cut locus for a given point is zero. On the standard round sphere, the cut locus of a point consists of the single point opposite of it. The distance function $r_p(x)$ is differentiable everywhere except at the point $p$ and its cut locus $\cut(p)$.

\section{Appendix. Connection Laplacian and Functional Derivative on Vector Fields}
In this section, we introduce the connection Laplacian operator and the functional derivative on vector fields.

\subsection{Connection Laplacian}
Let $\Gamma(T\mathcal{M})$ denote the space of smooth vector fields. Let $X, Y\in \Gamma(T\mathcal{M})$ be two smooth vector fields. Then it can be easily verified that $X+Y$ and $fX$ are smooth vector fields for any smooth function $f$. Therefore $\Gamma(T\mathcal{M})$ is a vector space. Let $T$ be a $(1, 1)$-tensor. Then $T$ is a linear operator $T: \Gamma(T\cM) \rightarrow \Gamma(T\cM)$. The adjoint $T^*$ can be defined as a linear operator satisfying
\[
g(TX, Y) = g(X, T^*Y),
\]
for all vector fields $X, Y\in \Gamma(T\cM)$. We can further define the inner product of two $(1,1)$ tensors $T_1, T_2$ as
\[
\langle T_1, T_2 \rangle := \tr(T_1^* T_2),
\]
where $\tr$ denotes the trace. The trace of a linear operator $L :  \Gamma(T\cM) \rightarrow \Gamma(T\cM)$ can be given as $\tr(L) = \sum_{i} g(L(\partial_i), \partial_i)/g(\partial_i, \partial_i)$ where $\{ \partial_i \}$ is a basis of $\Gamma(T\cM)$.

Note that the connection $\nabla$ is a linear operator:
\[
\nabla : \Gamma(T\cM) \rightarrow L(\Gamma(T\cM), \Gamma(T\cM)),
\]
where $L(\Gamma(T\cM), \Gamma(T\cM))$ denotes the space of linear operators on $\Gamma(T\cM)$. We can get an adjoint operator $\nabla^*: L(\Gamma(T\cM), \Gamma(T\cM)) \rightarrow \Gamma(T\cM)$ defined implicitly by
\[
\int_{\cM} g(\nabla^*T, Y) = \int_{M}\langle T, \nabla Y \rangle,
\]
for any $X, Y\in \Gamma(T\cM)$. If we set $T = \nabla X$, then we get a linear operator $\nabla^*\nabla: X \mapsto \nabla^*\nabla X$. $\nabla^*\nabla$ is called the \emph{connection Laplacian} on vector fields. By the construction of the connection Laplacian, it is not difficult to see that it is a positive semidefinite self-adjoint operator. By definition, we also have the following equation:
\[
\int_{\cM} g(\nabla^*\nabla X, Y) = \int_{M}\langle \nabla X, \nabla Y \rangle.
\]

There is another way to define the connection Laplacian using local coordinates. Consider the second covariant derivative and take the trace $\sum_{i=1}^d \nabla^2_{\partial_i, \partial_i} X$ with respect to some orthonormal frame $\partial_i$. This can be seen to be invariantly defined. We shall use the notation:
\begin{equation}
\begin{aligned}
\tr(\nabla^2 X)  &= \sum_{i=1}^d \nabla^2_{\partial_i, \partial_i} X,\\
\tr(\nabla^2 )  &= \sum_{i=1}^d \nabla^2_{\partial_i, \partial_i}.
\end{aligned}
\end{equation}
The connection Laplaican $\nabla^*\nabla$ equals to $-\tr(\nabla^2)$~\cite{RiemannianGeometry}.

\subsection{Functional Derivative}
Our objective function for learning a vector field $V$ is defined as follows:
\begin{equation}\label{eq:vf-continuous-obj-v2}
\min_{V} E(V) := \int_{\mathcal{M}} \| V - V^0\|^2  + t \int_{\mathcal{M}} \| \nabla V\|_{\mathrm{HS}}^2,
\end{equation}
where $\|\cdot\|_{\HS}$ denotes the Hilbert-Schmidt tensor norm \cite{Defant1993tensor}. We already showed in the last section that Eq.~\eqref{eq:vf-continuous-obj-v2} is equivalent to the following equation:
\[
\min_{V} E(V) := \int_{\mathcal{M}} \| V - V^0\|^2  + t \int_{\mathcal{M}} g (V, \nabla^*\nabla V ).
\]
The necessary condition of $E(V)$ to have an extremum at $V$ is that the functional derivative ${\delta E(V)}/{\delta V} = 0$~\cite{Abraham:1988:manifolds}. Next we show how to compute the functional derivative ${\delta E(V)}/{\delta V}$.

\begin{Def}[\cite{Abraham:1988:manifolds}]
Let $M, N$ be normed vector spaces, $U$ an open subset of $M$ and let $f:U \subset M \rightarrow N$ be a given mapping. Let $u_0 \in U$. We say that $f$ is differentiable at the point $u_0$ provided there is a bounded linear mapping $Df(u_0): M \rightarrow N$ such that for every $\epsilon> 0$, there is a $\delta > 0$ such that whenever $0 < \| u-u_0 \|<\delta$, we have
\[
\frac{ \| f(u) - f(u_0) - Df(u_0) \cdot (u-u_0)  \| }{ \| u - u_0 \| } < \epsilon,
\]
where $ \| \cdot \| $ represents the norm on the appropriate space and where the evaluation of $Df(u_0)$ on $e\in M$ is denoted $Df(u_0)\cdot e$.
\end{Def}
The derivative $Df(u_0)$ if it exist, is unique~\cite{Abraham:1988:manifolds}. $f:U \subset M \rightarrow N$ is differentiable at $u_0\in U$ if and only if there exists a linear mapping $Df(u_0) \in L(M, N)$ such that
\[
f(u_0+e) = f(u_0) + Df(u_0)\cdot e + o(e),
\]
where $o(e)$ denotes a continuous function of $e$ defined in a neighborhood of the origin of a normed vector space $M$, satisfying $\lim_{e\rightarrow 0} ( o(e)/\|e\| ) = 0$.

For any two vector fields $X$ and $Y$, define the inner product $(\cdot, \cdot)$ on the space of vector fields as
\[
(X, Y) = \int_{\mathcal{M}} g(X, Y) dx.
\]
The norm of $X$ can be defined as $\|X\|^2 = \int_{\mathcal{M}} g(X, X) dx$. Therefore $(\Gamma(T\cM), \| \cdot \|)$ is a normed vector space. We first consider the functional $(V -  V^0, V -  V^0) $. We have
\[
\begin{aligned}
&(V -  V^0 + e, V -  V^0 + e) \\&= (V -  V^0, V -  V^0) + 2(V -  V^0, e) + (e, e).
\end{aligned}
\]
Therefore $2(V - V^0)$ is a derivative of $(V -  V^0, V -  V^0) $ with respect to $V$. Since if the derivative exists, it is unique. Therefore $2(V - V^0)$ is exactly the unique derivative of $(V -  V^0, V -  V^0) $. Similarly, we expand the functional $(V, \nabla^*\nabla V)$ as
\[
\begin{aligned}
&(V + e, \nabla^*\nabla (V + e)) \\
&= (V , \nabla^*\nabla V) + (V, \nabla^*\nabla e) + (e, \nabla^*\nabla V) + (e, \nabla^*\nabla e) \\
&=  (V , \nabla^*\nabla V) + 2(e, \nabla^*\nabla V) +  (e, \nabla^*\nabla e) \\
&=  (V , \nabla^*\nabla V) + 2(e, \nabla^*\nabla V) +  o(e)
\end{aligned}
\]
where the second equality holds due to that $\nabla^*\nabla$ is a self-adjoint operator and the third equality holds due to that $\nabla^*\nabla$ is a bounded operator. Therefore, the derivative of $(V, \nabla^*\nabla V)$ with respect to $V$ is $2\nabla^*\nabla V$. In summary, we have
\[
\frac{\delta E(V)}{\delta V} = 2V - 2V^0 + 2t\nabla^*\nabla V.
\]

\section{Appendix. Proof of Main Theorems}\label{sec:distance-function-vector-field}

\subsection{Proof of Theorem~\ref{Them:distance_function_vector_field_equivalence}}
\begin{proof}
($\Rightarrow$)
(a) Since $\mathcal{M}$ is a complete manifold, according to Hopf-Rinow Theorem~\cite{RiemannianGeometry}, it is also geodesic complete. That is, for every $p$ in $\mathcal{M}$, the exponential map $\exp_p$ is defined on the entire tangent space $T_{p}\mathcal{M}$. Therefore, for any $x\in \mathcal{M}\setminus p\cup \cut(p)$, there exist some $v \in T_p\mathcal{M}$ such that $x = \exp_p(v)$. Actually $\exp_p(tv), 0 \leq t \leq 1$, is the unique minimal geodesic connecting $p$ and $x$. Thus $r(x) = d(p, x) = l(\exp_p(v)) = \|v\| = \|\exp_p^{-1}(x)\|$.

(b) Since the distance function $d(p, x)$ has constant speed 1, we have $\| \partial_r \| =1$. For any $q\in \mathcal{M}$, let $\gamma(s): [0, t] \rightarrow \mathcal{M}$ be a curve connecting $p$ and $q$, i.e., $\gamma(0) = p$ and $\gamma(t)=q$. Then the length of the curve satisfies the following inequality:
\begin{eqnarray*}
l(\gamma) &=& \int_{0}^t \|\gamma'(s)\|ds \\
&=& \int_{0}^t \|\partial_r\| \| \gamma'(s)\|ds \\
&\geq & \int_{0}^t |g(\partial_r, \gamma'(s))| ds \\
&\geq& | \int_{0}^t \frac{d r\circ \gamma}{ds}ds | \\
&=& r(q)-r(p)=r(q).
\end{eqnarray*}
Here the second equality holds due to $\|\partial_r\|=1$ and first inequality holds due to the Cauchy-Schwarz inequality. Since $\gamma$ is arbitrary, we have $d(p, q) \geq r(q)$. Let $\gamma$ be an integral curve~\cite{LEESmoothManifold} of $\partial_r$, i.e., $\gamma'(s) = \partial_r \circ \gamma(s)$. Then the equality holds in the Cauchy-Schwarz inequality and $\frac{d r\circ \gamma}{ds} >0$. Thus $l(\gamma) = |r(q) - r(p)| = r(q) = d(p, q)$. This shows that the integral curve of $\partial_r$ must be a minimal geodesic. According to the definition of geodesic, we have $\nabla_{\partial_r} \partial_r = 0$ hold whenever $r$ is smooth.

($\Leftarrow$)
We first show that each integral curve of $\partial_r$ in $\mathcal{M}\setminus p \cup \cut(p)$ must be a geodesic. Let $\gamma(s)$ be an integral curve of $\partial_r$. According to the definition of the integral curve, we have $\gamma'(s) = \partial_r$. Therefore $\nabla_{\gamma'(s)}\gamma'(s) = \nabla_{\partial_r}\partial_r = 0$ which implies $\gamma(s)$ is a geodesic. Since geodesics have constant nonzero speed, we have $\|\gamma'(s)\| = \|\partial_r\| >0$ for each integral curve $\gamma(s)$. Recall that if a vector field is smooth at some point, then there exists a unique integral curve passing through it~\cite{LEESmoothManifold}. We have $\|\gamma'(s)\| = \|\partial_r\| >0$ holds whenever $r$ is smooth.

Next we show each integral curve $\gamma(s)$ must pass through $p$. We first prove this claim in the neighborhood $U$ of $p$. For any point $x\in U$, there exists some $t_0 \geq 0$ and a unit vector $v\in T_p\mathcal{M}$ such that $x = \exp_p(t_0v)$. Since $r(x) = \|\exp_p^{-1}(x)\|$, according to the Gauss lemma~\cite{RiemannianGeometry}, we have $\partial_r = D\exp_p(\partial_r)$. Thus the integral curve of $\partial_r$ passing through $x$ is the exponential map $\exp_p(tv)$, and we have $\|\partial_r\| = \|D_t\exp_p(tv)\| = \|v\| =1$. So far, for any point $x\in U$, we have proved that the integral curve passing through $x$ must be a geodesic passing through $p$ and vice versa. Next we show for any point $x \in \mathcal{M}\setminus p\cup \cut(p)$, the integral curve passing through $x$ must pass through $p$. Since the manifold is complete, according to Hopf-Rinow Theorem~\cite{RiemannianGeometry}, there always exists a minimal geodesic connecting any two points on the manifold. Let $\gamma(s)$ be the unique minimal geodesic connecting $p$ and $x$ satisfies $\gamma(0) = p$ and $\gamma(t_1) = x$. Then $\gamma(s)\cap U$ is an integral curve of $\partial_r$ in $U$. Note that each integral curve of $\partial_r$ must be a geodesic, due to the global uniqueness of geodesic~\cite[please see lemma. 7 of chap. 5]{RiemannianGeometry}, so $\gamma(s)$ is an integral curve of $\partial_r$ on $\mathcal{M}\setminus p\cup \cut(p)$.

Now we are ready to prove $r$ is a distance function, i.e., $r(x) = d(p, x)$. For each point $x\in \mathcal{M}\setminus p \cup \cut(p)$, we show that $r(x) = d(p, x)$. Let $\gamma(s)$ be the integral curve connecting $p$ and $x$, without loss of generality, we assume that $\gamma(0)=p, \gamma(t)=x$ for some $t>0$. According to the condition (a), we have $\|\gamma'(s)\| = \|\partial_r\| = 1$ for small $s$. Note that $\gamma$ is also a geodesic, thus $\gamma$ has constant speed which implies $\|\gamma'(s)\| = \|\partial_r\| = 1$ for $s\in [0,t]$. Thus
$$
r(x) = r(x) - r(p) = \int_{p}^x \|\partial_r\| dx = \int_{0}^t \| \gamma'(s) \| ds = t.
$$
Since $l$ is a geodesic with speed 1, we have $d(p, x) = \int_{0}^t \| \gamma'(s) \| ds = t$ which implies $r(x) = d(p,x)$.

Finally, we show that $r(x) = d(p, x)$ holds even if $x\in \mathrm{Cut}(p)$. Choose any geodesic $\gamma$ connecting $p$ and $x$. We first show that there is no any other cut point on $\gamma$. If not, assume there is a point $z$ on $\gamma$ between $p$ and $x$. Then we have $d(p,z) < d(p,x)$ which is contradict to the definition of the cut point. Since $x$ is the unique cut point on $\gamma$, we can choose a sequence of points $x_i$ on $\gamma$ such that its limit being $x$. Then we have $r(x) = \lim_{x_i \rightarrow x} r(x_i)= \lim_{x_i \rightarrow x} d(p, x_i) = d(p, x)$.
\end{proof}

\subsection{Proof of Theorem~\ref{Them:distance_function_unit_norm}}

\begin{proof}
($\Rightarrow$) is true throughout the proof of Theorem \ref{Them:distance_function_vector_field_equivalence}. Next we show ($\Leftarrow$). Compared to Theorem~\ref{Them:distance_function_vector_field_equivalence}, we only have to show $\nabla _{\partial_r} \partial_r = 0$ on $\mathcal{M} \setminus p\cup \mathrm{Cut}(p)$. Let $S(\cdot)$ denote the (1,1) version of the Hessian of $r$, $S(\cdot) = \nabla_{\cdot}\partial r$, i.e., $\hess r(X, Y) = g(S(X), Y)$. It is evident that $\hess r$ is a symmetric tensor, thus $\hess r(X, Y) = \hess r(Y, X)$.

For any vector field $Y$ on $\mathcal{M} \setminus p\cup \mathrm{Cut}(p)$, then
\begin{eqnarray*}
g(\nabla_{\partial_r} \partial_r , Y) &=& \mathrm{Hess} r (\partial_r, Y) \\
&=&  \mathrm{Hess} r (Y, \partial_r) \\
&=& g(\nabla_Y \partial_r, \partial_r) \\
&=& \frac{1}{2} Yg(\partial_r, \partial_r) \\
&=& \frac{1}{2} Y(1) = 0.
\end{eqnarray*}
The second equality holds due to the symmetry of $\hess r$, the third equality holds due to the definition of $\hess r$ and the fourth equality holds due to the property of the covariant derivative. Thus $\nabla_{\partial_r} \partial_r = 0$ holds on $\mathcal{M} \setminus p\cup \mathrm{Cut}(p)$. Since $r$ satisfies conditions (a) and (b) in Theorem \ref{Them:distance_function_vector_field_equivalence}, $r$ is a distance function based at $p$.
\end{proof}

{
\bibliographystyle{abbrv}
\bibliography{Reference_dcai}
}

\end{document}